\documentclass[conference]{IEEEtran}
\usepackage{amsmath, amssymb}
\usepackage{graphicx}
\usepackage{booktabs}
\usepackage{tikz}
\usetikzlibrary{arrows.meta,positioning}
\usepackage[dvipsnames]{xcolor}
\usepackage{threeparttable}
\usepackage{hyperref}
\usepackage{cite}
\usepackage{algpseudocode}
\usepackage[T1]{fontenc}
\usepackage{lmodern}
\usepackage[utf8]{inputenc}
\usepackage{float} 
\usepackage{multirow}
\usepackage[caption=false,font=footnotesize]{subfig}

\title{Graph-Coarsening Approach for the Capacitated Vehicle Routing Problem with Time Windows}

\author{
    \IEEEauthorblockN{Mustafa Mert Özyılmaz}
    \IEEEauthorblockA{
        Sorbonne Université \\
        Email: mustafa\_mert.ozyilmaz@etu.sorbonne-universite.fr
    }
}

\begin{document}
\maketitle

\begin{abstract}
The Capacitated Vehicle Routing Problem with Time Windows (CVRPTW) is a fundamental NP-hard optimization problem in logistics. Solving large-scale instances remains computationally challenging for exact solvers. This paper introduces a multilevel graph coarsening and refinement strategy that aggregates customers into meta-nodes based on a spatio-temporal distance metric.  The reduced problem is solved using both classical heuristics and quantum annealing hardware, then expanded back into the original space with arrival times recomputed and constraint violations recorded. Comprehensive experiments on Solomon benchmarks demonstrate that our method significantly reduces computation time while preserving solution quality for classical heuristics. For quantum solvers, experiments across all 56 Solomon instances at $N=5$ and $N=10$ customers show that coarsening consistently reduces computation time and, on clustered (C-type) instances, simultaneously reduces vehicle count and route duration with no feasibility loss.  Coarsening effectiveness is strongly instance-structure dependent: C-type instances achieve 100\% post-coarsening feasibility with measurable quality improvements, while narrow-window random (R-type) instances present structural constraints that limit achievable coarsening depth.
\end{abstract} 
\section{Introduction}
Vehicle Routing Problems (VRPs) are central to logistics optimization and remain computationally intractable at realistic scales. A critical variant is the Capacitated VRP with Time Windows (CVRPTW), where both capacity limits and customer time windows must be respected. Traditional heuristics and metaheuristics provide approximate solutions but struggle on very large instances.

Recent advances in quantum optimization and graph simplification motivate new preprocessing strategies. In this work, we investigate spatio-temporal graph coarsening as a means to accelerate CVRPTW solvers.
\footnote{Code, full experimental results, and route visualizations are provided at \url{https://github.com/mkingmking/graph-coarsening}.}

\textbf{Our main contributions are:}
\begin{itemize}
  \item A spatio-temporal distance metric integrating travel times and time windows for node aggregation.  
  \item A feasibility-aware coarsening routine that enforces merge compatibility with respect to time windows and capacity, combined with reversible inflation that recomputes arrival times to detect and record constraint violations.  
  \item Experimental evaluation on Solomon benchmarks demonstrating efficiency gains and preserved solution quality.  
\end{itemize}

\section{Related Work}

Over the past decades, intensive research has been conducted on harnessing quantum computing for solving complex optimization problems and achieving faster computation. Early quantum algorithms such as Grover’s search~\cite{grover1996} and Shor’s factoring~\cite{shor1997} demonstrated the potential for significant speedups on specific tasks, sparking interest in quantum approaches to broader problem classes. One prominent paradigm is \emph{adiabatic quantum computing}, implemented via quantum annealing (QA) on specialized hardware~\cite{kadowaki1998,farhi2000}. Quantum annealing exploits quantum tunneling effects to approximate the ground state of an Ising-model Hamiltonian encoding a given optimization problem. In this framework, many NP-hard problems are formulated in either a \emph{Ising model} or an equivalent \emph{Quadratic Unconstrained Binary Optimization (QUBO)} form, an energy function over binary variables whose minimization corresponds to the objective of the problem~\cite{kochenberger2014}. While the general QUBO problem is itself NP-hard, its strength lies in its versatility: a wide range of combinatorial optimization tasks can be reduced to these formulations. \\ 


Early approaches to solving the VRP problem include the Clarke-Wright savings heuristic~\cite{clarke1964}, which quickly became one of the most popular constructive techniques. Numerous meta-heuristics and exact algorithms followed, but VRP instances of practical size often remain computationally intractable.\\  An important variant is the \emph{Capacitated VRP with Time Windows (CVRPTW)}, which imposes both vehicle capacity limits and specific time windows for service. These constraints couple routing with scheduling, increasing complexity. Qi et al.~\cite{qi2011} introduced a spatiotemporal distance metric for CVRPTW, integrating spatial and temporal considerations. 

In recent years, there has been growing exploration of \emph{quantum and hybrid quantum-classical algorithms} for VRP and its variants. For example, Borowski et al.~\cite{borowski2020} developed hybrid algorithms for VRP and Capacitated VRP, including the Full QUBO Solver (FQS) and the Average Partitioning Solver (APS). These methods combine classical heuristics with D-Wave quantum annealers, illustrating the viability of hybrid quantum-classical approaches. 

A complementary strategy is to reduce problem size via graph simplification. Nałęcz-Charkiewicz et al.~\cite{nalecz2025} proposed a \emph{graph coarsening approach} for VRP, pruning long-distance edges while preserving core connectivity. Their results showed a roughly 50\%  improvement in solution quality compared to using the same hybrid solvers without graph coarsening.

For CVRPTW, quantum-based approaches are emerging. Holliday et al.~\cite{holliday2025} studied an advanced QA method using D-Wave’s Constrained Quadratic Model (CQM) solver, achieving solutions with an average optimality gap of 3.86\%. The CQM framework extends binary quadratic models by explicitly incorporating linear and quadratic constraints over binary, integer, or continuous variables, allowing richer and more realistic problem formulations.\cite{benson2023cqm} Their hybrid approach combined quantum annealing with classical repair heuristics to enforce time windows. 
\\ \\
In this work, we build on these advances by extending the graph-coarsening framework of Nałęcz-Charkiewicz et al.~\cite{nalecz2025} to the Capacitated Vehicle Routing Problem with Time Windows (CVRPTW). Our variant explicitly integrates temporal feasibility by incorporating service durations, time windows, and a spatio–temporal distance metric during the merge process. We ensure temporal feasibility when pruning edges, and then solve the reduced problem using greedy \cite{cormen2009introduction} and savings heuristics~\cite{clarke1964}, as well as QUBO-based solvers including FQS and APS~\cite{borowski2020}. After inflation to the original graph, arrival times are recomputed and any residual constraint violations are recorded as solution metrics. Empirical results on benchmark CVRPTW instances demonstrate that our method improves solution quality, highlighting the potential of graph coarsening as a pre-processing step for quantum-enhanced optimization.

\section{Problem Formulation}

\subsection{Mathematical Definition of the CVRPTW}

The Capacitated Vehicle Routing Problem with Time Windows (CVRPTW) is a well-known NP-hard combinatorial optimization problem in logistics~\cite{holliday2025}. It is defined on a graph 
\(G = (V, A)\), where \(V = \{v_0, v_1, \ldots, v_n\}\) is the set of nodes (vertices) and \(A\) is the set of arcs (edges) connecting them. Node \(v_0\) represents the \emph{depot} (start and end location for vehicles), and nodes \(v_1, \dots, v_n\) represent \emph{customers}~\cite{borowski2020}. Each customer node \(v_i\) has an associated demand \(d_i\) (the amount of goods to deliver or pick up), a service time \(s_i\) (the time required to serve that customer), and a time window \([e_i,\, l_i]\) specifying the earliest and latest permissible start times for service at that customer~\cite{qi2011}. We are given a fleet of \(K\) identical vehicles, each with capacity \(Q\), all initially located at the depot~\cite{holliday2025}. A travel cost or distance \(c_{ij}\) is defined for each arc \((i,j) \in A\), representing the cost or travel time from node \(i\) to node \(j\)~\cite{borowski2020}. 

A \emph{route} (or tour) for a single vehicle is a sequence of nodes starting and ending at the depot node \(v_0\)~\cite{borowski2020}. A feasible solution to the CVRPTW consists of a set of up to \(K\) routes such that every customer node is included in exactly one route and all operational constraints are satisfied. Formally, we can introduce binary decision variables \(x_{i,j,k} \in \{0,1\}\) equal to 1 if and only if vehicle \(k\) travels directly from node \(i\) to node \(j\) (and 0 otherwise)~\cite{holliday2025}. We may also introduce time variables \(t_{i}\) to denote the arrival (or service start) time at node \(i\) for whichever vehicle services customer \(i\). The CVRPTW can then be formulated as an integer programming model. The \emph{objective} is to minimize the total travel cost (e.g., total distance or time) of all routes~\cite{holliday2025}:

\[
 \min \; \sum_{k=1}^K \sum_{i \in V} \sum_{j \in V} c_{ij}\, x_{i,j,k},
\] 

subject to the following constraints that define feasible vehicle routes:

\begin{itemize}
  \item \textbf{Depot start/end:} Each vehicle route begins and ends at the depot. This means every vehicle \(k\) leaves the depot \(v_0\) at most once and returns to the depot at the end of its route~\cite{borowski2020,holliday2025}. If a vehicle is not used, it simply stays at the depot.
  \item \textbf{Visit each customer exactly once:} Every customer must be visited by exactly one vehicle. Each customer node \(v_i (i\ge 1)\) has exactly one incoming arc and one outgoing arc in the set of chosen routes~\cite{holliday2025}. This ensures that all customers are served and no customer is serviced twice.
  \item \textbf{Capacity constraint:} The total demand served by any single route cannot exceed the vehicle capacity \(Q\). In the Solomon benchmark instances used in this work the fleet is homogeneous, with each vehicle having identical capacity \(Q\) = 200. Equivalently, for each vehicle \(k\), the sum of demands of customers on its route must satisfy \(\sum_{i \text{ on route } k} d_i \le Q\)~\cite{holliday2025}.
  \item \textbf{Time window constraints:} Each vehicle must arrive within the specified time window of each customer. If \(t_i\) is the arrival (start-of-service) time at customer \(i\), we require \(e_i \le t_i \le l_i\) for all customers \(i\)~\cite{holliday2025}. Arriving before the window opens (\(t_i < e_i\)) incurs waiting until \(e_i\), and arriving after the latest time \(l_i\) is not allowed. In addition, if a vehicle travels from customer \(i\) to \(j\) consecutively on its route, the arrival time must satisfy \(t_j \ge t_i + s_i + \text{travel\_time}_{ij}\). These timing constraints ensure no route violates any customer’s scheduling window~\cite{borowski2020}.
\end{itemize}

This formulation is subject to the standard VRPTW constraints: 
each customer must be visited exactly once, vehicle flows must 
be conserved, every route must start and end at the depot, 
vehicle capacity limits must be respected, and service must occur 
within customer time windows. Additional linking constraints 
propagate travel and service times along the route and eliminate 
subtours.

All decision variables \(x_{i,j,k}\) are binary and the time variables \(t_i\) are continuous. This formulation is a mixed-integer linear program (MILP) describing the CVRPTW. It captures the requirement that each customer is served exactly once and respects vehicle capacities and time scheduling restrictions~\cite{holliday2025}. The classical objective is to minimize the total distance or travel cost of all routes~\cite{borowski2020}, though in some variations one might also consider minimizing the number of vehicles used or other criteria.

\subsection{QUBO Formulation for CVRPTW}

To leverage quantum or quantum-inspired optimization methods, the CVRPTW can be reformulated as a \emph{Quadratic Unconstrained Binary Optimization (QUBO)} problem. A QUBO formulation encodes the routing decision variables and relevant constraints into a quadratic objective function defined over binary variables, without explicit linear constraints~\cite{kochenberger2014}. In a QUBO model, all constraints must be incorporated into the objective via penalty terms, since the formulation does not allow separate constraints~\cite{papalitsas2019}. 

This means we introduce penalty coefficients that heavily penalize any violation of the original CVRPTW constraints. For example, a term 
\(P \cdot ( \sum_{k,j} x_{i,j,k} - 1)^2\) 
can be added to the QUBO objective for each customer \(i\) to enforce that exactly one vehicle visits customer \(i\)~\cite{holliday2025}. Similarly, capacity constraints \(\sum_{i}\sum_{j} d_i\,x_{i,j,k} \le Q\) can be imposed by adding penalties for routes that exceed capacity, and time window constraints can be encoded by penalizing any scheduling that falls outside the allowed window~\cite{borowski2020}. Because time window and capacity restrictions are originally inequality constraints, they are particularly challenging to handle in a QUBO model~\cite{holliday2025}. A common approach is to introduce additional slack binary variables to transform each inequality into an equality, which can then be translated into a quadratic penalty term~\cite{papalitsas2019}. This transformation increases the number of binary variables but guarantees that all constraints are expressed consistently within the quadratic (degree-2) framework of QUBO.

In summary, the QUBO representation of the CVRPTW involves constructing a single quadratic objective function \(H(x)\) such that:

\begin{itemize}
  \item \(H(x)\) includes a term for the original routing objective (total distance) in binary form, and  
  \item \(H(x)\) includes large-weight penalty terms for each constraint (visiting each customer once, capacity limit, time windows, depot start/return, etc.).
\end{itemize}

The result is a quadratic binary optimization problem whose ground-state solution corresponds to an optimal or near-optimal set of routes for the CVRPTW~\cite{papalitsas2019}. Prior works have demonstrated how to construct such QUBO models for vehicle routing problems. For instance, Papalitsas et al.~\cite{papalitsas2019} developed a QUBO model for the Traveling Salesman Problem with Time Windows, encoding time window constraints with binary variables and penalties. Likewise, recent studies have formulated capacitated VRPs as QUBO by embedding route feasibility conditions into the objective function~\cite{borowski2020}. Although formulating the CVRPTW as a QUBO is complex - especially due to the time window inequalities - it enables the use of quantum annealers and digital annealing hardware to search for high-quality solutions by minimizing the combined objective~\cite{holliday2025}.

\section{Proposed Method}

Our approach extends the classical graph coarsening framework for VRP by explicitly integrating temporal constraints required by CVRPTW. In particular, the coarsening process is adapted to account for service durations and customer time windows, ensuring that aggregated super-nodes remain feasible with respect to both spatial and temporal requirements. The method is characterized by five main elements:
\begin{itemize}
  \item augmentation of vertices with service time and time window attributes,
  \item definition of a combined spatio--temporal distance metric,
  \item explicit enforcement of temporal compatibility during node merges,
  \item propagation of aggregated time windows through successive coarsening layers, and
  \item reversible inflation to reconstruct feasible timed routes in the original graph.
\end{itemize}

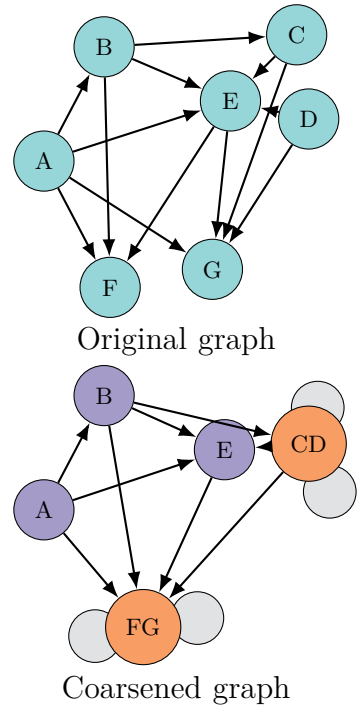
\begin{figure}[htbp]
\centering
\begin{tikzpicture}[scale=0.8,
  vertexA/.style={circle, draw, fill=TealBlue!45, minimum size=8mm, inner sep=0pt, font=\small},
  vertexB/.style={circle, draw, fill=BlueViolet!40, minimum size=8mm, inner sep=0pt, font=\small},
  merged/.style={circle, draw, fill=Orange!75, minimum size=10mm, inner sep=0pt, font=\small},
  ghost/.style={circle, draw, fill=Gray!25, minimum size=7mm, inner sep=0pt},
  edge/.style={-Latex, line width=0.8pt},
  weak/.style={-Latex, draw=gray!60, line width=0.4pt}
]

\begin{scope}[yshift=5.8cm]
  \node[vertexA] (A) at (-2.2,0.2) {A};
  \node[vertexA] (B) at (-1.2,2.1) {B};
  \node[vertexA] (C) at (2.0,2.3) {C};
  \node[vertexA] (D) at (2.2,0.9) {D};
  \node[vertexA] (E) at (0.9,1.2) {E};
  \node[vertexA] (F) at (-1.1,-1.9) {F};
  \node[vertexA] (G) at (0.6,-1.6) {G};

  \draw[edge] (A) -- (F);
  \draw[edge] (B) -- (C);
  \draw[edge] (E) -- (G);
  \draw[edge] (C) -- (E);
  \draw[edge] (D) -- (E);

  \draw[edge] (A) -- (B);
  \draw[edge] (A) -- (E);
  \draw[edge] (A) -- (G);
  \draw[edge] (B) -- (E);
  \draw[edge] (B) -- (F);
  \draw[edge] (E) -- (F);
  \draw[edge] (C) -- (G);
  \draw[edge] (D) -- (G);

  \node[font=\large] at (0,-2.8) {Original graph};
\end{scope}

\begin{scope}
  \node[vertexB] (A2) at (-2.2,0.2) {A};
  \node[vertexB] (B2) at (-1.2,2.1) {B};
  \node[vertexB] (E2) at (0.8,1.2) {E};

  \node[ghost] at (2.35,1.9) {};
  \node[ghost] at (2.55,0.5) {};
  \node[merged] (CD) at (2.2,1.3) {CD};

  \node[ghost] at (-1.35,-1.9) {};
  \node[ghost] at (0.35,-1.6) {};
  \node[merged] (FG) at (-0.55,-1.75) {FG};

  \draw[edge] (A2) -- (FG);
  \draw[edge] (B2) -- (CD);
  \draw[edge] (E2) -- (FG);
  \draw[edge] (CD) -- (E2);

  \draw[edge] (B2) -- (FG);
  \draw[edge] (B2) -- (E2);
  \draw[edge] (CD) -- (FG);
  
  \draw[edge] (A2) -- (B2);
  \draw[edge] (A2) -- (E2);

  \node[font=\large] at (0,-2.8) {Coarsened graph};
\end{scope}

\end{tikzpicture}
\caption{Illustration of the graph coarsening procedure. Nodes $C$ and $D$ are merged into super-node $CD$, and nodes $F$ and $G$ into super-node $FG$.}
\label{fig:graph-coarsening}
\end{figure}

\begin{figure}[t]
\centering
\begin{tikzpicture}[
    node distance=1.2cm,
    process/.style={rectangle, draw=black, rounded corners, fill=blue!10, text width=7.5cm, align=center, minimum height=1.0cm},
    io/.style={rectangle, draw=black, fill=gray!15, text width=7.5cm, align=center, minimum height=1.0cm},
    arrow/.style={thick,->,>=latex},
    every node/.style={font=\small}
]

\node[io] (input) {Input CVRPTW Graph \\ $G=(V,E)$ with customer coordinates, service durations, and time windows};
\node[process, below=of input] (augment) {Graph Augmentation \\ Add service time $s_i$, time window $[e_i,\ell_i]$, and nominal visit time $t_i$ to each node; define travel-time function $\tau_{ij}(t)$};
\node[process, below=of augment] (coarsen) {Spatio--Temporal Coarsening \\ Compute $D_{ij} = \alpha \tau_{ij} + \beta \Delta T_{ij}$; enforce merge feasibility; aggregate time windows $[e',\ell']$};
\node[process, below=of coarsen] (solver) {Optimization on Coarsened Graph $G'$ \\ Apply classical heuristic (Greedy or Savings) to compute a reduced solution};
\node[process, below=of solver] (inflate) {Inflation (Route Reconstruction) \\ Reverse merges from history $L$; expand super-nodes; restore full time-dependent route feasibility};
\node[io, below=of inflate] (output) {Output: Feasible, optimized routes on original graph $G$};

\draw[arrow] (input) -- (augment);
\draw[arrow] (augment) -- (coarsen);
\draw[arrow] (coarsen) -- (solver);
\draw[arrow] (solver) -- (inflate);
\draw[arrow] (inflate) -- (output);

\end{tikzpicture}
\caption{Overall workflow of the proposed spatio--temporal graph coarsening framework for CVRPTW. 
The pipeline proceeds top--down: from graph augmentation and coarsening, through reduced-graph optimization, to route inflation and feasibility restoration.}
\label{fig:methodology_vertical}
\end{figure}
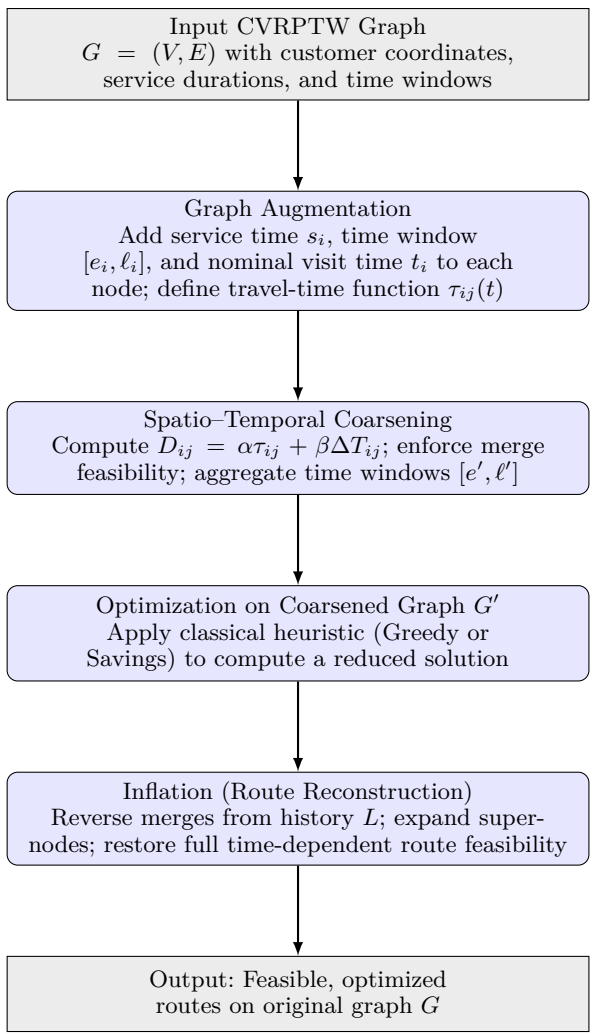

\subsection{Graph Model Augmentation}
Each customer node $i$ in the graph is described by spatial coordinates $(x_i,y_i)$, a service duration $s_i$, and a time window $[e_i,\ell_i]$. To facilitate temporal reasoning, we associate each node with a nominal visit time $t_i$, which by default is chosen as the midpoint
\[
t_i = \tfrac{e_i + (\ell_i - s_i)}{2}.
\]
Alternative definitions, such as the earliest feasible start time $t_i=e_i$ or an estimated arrival time obtained from a baseline tour, may also be employed. 

Edges $(i,j)$ are annotated with a travel time function $\tau_{ij}(t)$, which may be time-dependent. Within the coarsening routine, this is approximated either by evaluating at a representative time (e.g., midpoint of the merged time window) or, for safety, by using an upper bound.

\subsection{Temporal Separation}
To determine whether two vertices can be merged without violating time windows, we introduce a temporal separation measure. For vertices $i$ and $j$, the separation is defined as
\[
\Delta T_{ij} = \max\{0,\, e_j - (t_i + s_i + \tau_{ij})\}
\quad \text{or} \quad |t_i - t_j|,
\]
depending on whether strict feasibility checks or nominal alignment are preferred. This measure ensures that only pairs of customers whose schedules are mutually compatible are considered for merging.

\subsection{Spatio--Temporal Distance Metric}
The edge weight used during coarsening combines spatial and temporal components:
\[
D_{ij} = \alpha \cdot \tau_{ij} + \beta \cdot \Delta T_{ij},
\]
where $\tau_{ij}$ denotes travel time and $\Delta T_{ij}$ the temporal separation. The coefficients $\alpha$ and $\beta$ balance the influence of geographical proximity against scheduling alignment.

\section{Algorithm Overview}
We apply a multilevel coarsening of the CVRPTW graph, followed by route reconstruction (inflation). Feasibility is enforced at each merge by checking travel times in both possible service orders.

\subsection{Merge Feasibility}
For an edge $(i,j)$, feasibility is evaluated by verifying that at least one service order is possible:
\[
\begin{aligned}
  \mathrm{feas}_{i \to j} &= e_i \le \ell_j - s_j - \tau_{ij} - s_i, \\
  \mathrm{feas}_{j \to i} &= e_j \le \ell_i - s_i - \tau_{ji} - s_j.
\end{aligned}
\]
If both fail, the merge is disallowed. If feasible, the direction with greater slack is chosen:
\[
\begin{aligned}
  \mathrm{slack}_{i \to j} &= (\ell_j - s_j - \tau_{ij}) - (e_i + s_i), \\
  \mathrm{slack}_{j \to i} &= (\ell_i - s_i - \tau_{ji}) - (e_j + s_j).
\end{aligned}
\]

\subsection{Aggregated Time Window Propagation}
Once the service order $\pi$ (either $i \to j$ or $j \to i$) is fixed, a new aggregated window $[e',\ell']$ is computed. In the relaxed propagation, we assume (w.l.o.g.) the order is fixed to $i \to j$. A relaxed propagation scheme sets

\[
\begin{cases}
  e' = \min(e_i,\, e_j - (s_i + \tau_{ij})), \\
  \ell' = \max(\ell_j - s_j,\, \ell_i - (s_i + \tau_{ij})),
\end{cases}
\]
while a conservative alternative enforces
\[
e' = \max(e_i, e_j), \quad \ell' = \min(\ell_i, \ell_j).
\]

\subsection{Coarsening Routine}
The coarsening procedure iteratively merges nodes based on the smallest $D_{ij}$ values, subject to feasibility checks. A pseudocode outline is as follows:

\begin{algorithmic}[1]
\Require $G=(V,E)$ with attributes, depot $v_{0}$, rate $P$, radiusCoeff
\Ensure Coarsened graph $G'$, merge layers $L$
\State $G' \gets G$, $L \gets \emptyset$, $n_0 \gets |V|$
\While{$|V(G')| > P \cdot n_0$}
  \State Compute pairwise distances $D_{ij}$ for all $(i,j)\in E(G')$
  \State Sort edges by increasing $D_{ij}$
  \State Determine threshold $\rho$ from radiusCoeff
  \State Initialize $M \gets \emptyset$, $U \gets \{v_{0}\}$ \Comment{depot never merges}
  \ForAll{$(i,j)$ with $D_{ij} \le \rho$}
    \If{$i \in U$ or $j \in U$} \textbf{continue} \EndIf
    \State Check feasibility and slack; if infeasible \textbf{continue}
    \State Compute aggregated time window $[e',\ell']$ under order $\pi$
    \State $M \gets M \cup \{(i,j,\pi,e',\ell')\}$; $U \gets U \cup \{i,j\}$
  \EndFor
  \If{$M = \emptyset$} \textbf{break} \EndIf
  \ForAll{$(i,j,\pi,e',\ell') \in M$}
    \State Create new super-node $v_{ij}$ positioned at midpoint of $i,j$
    \State Assign attributes: $s_{ij} \gets s_i + s_j$, $t_{ij} \gets \tfrac{e'+\ell'}{2}$, etc.
    \State Add $v_{ij}$ to $V(G')$
    \State Reconnect edges: for each neighbor $k$ of $i$ or $j$, add edge $(v_{ij},k)$ with updated weights \State Recompute travel time $\tau(v_{ij},k)$ conservatively$^{*}$
    \State Record merge $(v_{ij},i,j,\pi)$ in $L$
    \State Remove $i,j$ (and incident edges) from $G'$
  \EndFor
\EndWhile
\State \Return $G',L$
\end{algorithmic}

\begin{tablenotes}[para,flushleft]
\footnotesize
The parameter \texttt{radiusCoeff} is the \emph{coarsening radius coefficient}. It scales the radius used by the coarsener to identify candidate nodes for merging. The radius is calculated with respect to the maximum dimension of the graph, and \texttt{radiusCoeff} serves as a multiplier. A larger value of \texttt{radiusCoeff} increases the radius, which may result in larger clusters being formed. This parameter therefore controls the granularity of the coarsening step and influences both the quality of merges and the eventual solution space explored.\\

\noindent $^{*}$ \textit{Note:} By “conservatively,” we mean recomputing $\tau$ so that any route feasible in the coarsened graph $G'$ 
remains feasible when expanded back to the original graph $G$. This notion is conceptually similar to 
preservation principles in graph reduction and coarsening literature, where structural or spectral properties 
are maintained across levels of abstraction \cite{jin2018spectral,hashemi2024survey}. 
A fully conservative computation may require more sophisticated logic (e.g., incorporating worst-case detours or slack propagation) 
to guarantee feasibility under inflation.

\end{tablenotes}

\subsection{Inflation (Route Reconstruction)}
After optimization on the reduced graph $G'$, we expand the solution back:
\begin{enumerate}
  \item Traverse the merge history $L$ in reverse.
  \item For each $(v_{ij},i,j,\pi)$, locate $v_{ij}$ on the route.
  \item Replace $v_{ij}$ with the ordered pair $(i,j)$ or $(j,i)$ according to $\pi$.
  \item Recompute arrival times downstream to verify time-window feasibility. Any residual violations are recorded as solution metrics.
\end{enumerate}

End-to-end solution feasibility depends not only on the coarsening step but on the quality of the solution found by the solver on the reduced graph. As the results demonstrate, this distinction is instance-scale dependent: at $N=5$, coarsening frequently restores feasibility; at $N=10$, the solver's tendency to converge to constraint-violating local minima limits the feasibility benefit of coarsening.

\subsection{Time-Dependent Travel Times}
If travel times are time dependent, $\tau_{ij}(t)$ is approximated during coarsening either by a representative fixed time (e.g., midpoint of the merged window) or by an upper bound. During inflation, actual arrival times are used to recompute $\tau_{ij}(t)$, thereby restoring full temporal accuracy.

\section{Computational Experiments}

\subsection{Experimental Setup}
The objective of our computational experiments is to evaluate the performance of the proposed spatio-temporal graph coarsening algorithm. We seek to demonstrate its effectiveness in reducing the size of large-scale Capacitated Vehicle Routing Problem with Time Windows (CVRPTW) instances and to quantify the impact of this reduction on solution quality when paired with classical VRP heuristics. All experiments were conducted on a single machine with a 8-core processor and 16 GB of RAM, running the MacOS operating system. The code, written in Python~3.9, relies on standard libraries such as \texttt{numpy}, \texttt{pandas}, and \texttt{scipy} for data handling and analysis, as well as \texttt{seaborn} and \texttt{matplotlib} for data visualization.


\begin{figure}[htbp]
\centering

\subfloat[C101]{%
  \includegraphics[width=0.48\columnwidth]{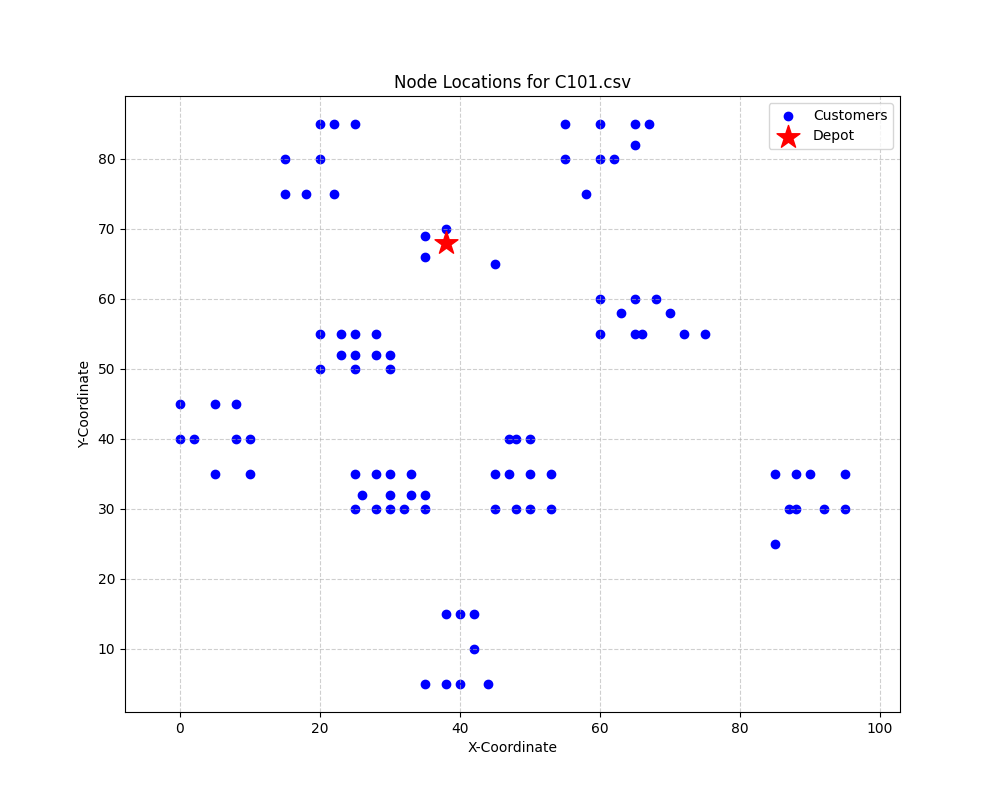}%
  \label{fig:c101}}%
\hfill
\subfloat[C103]{%
  \includegraphics[width=0.48\columnwidth]{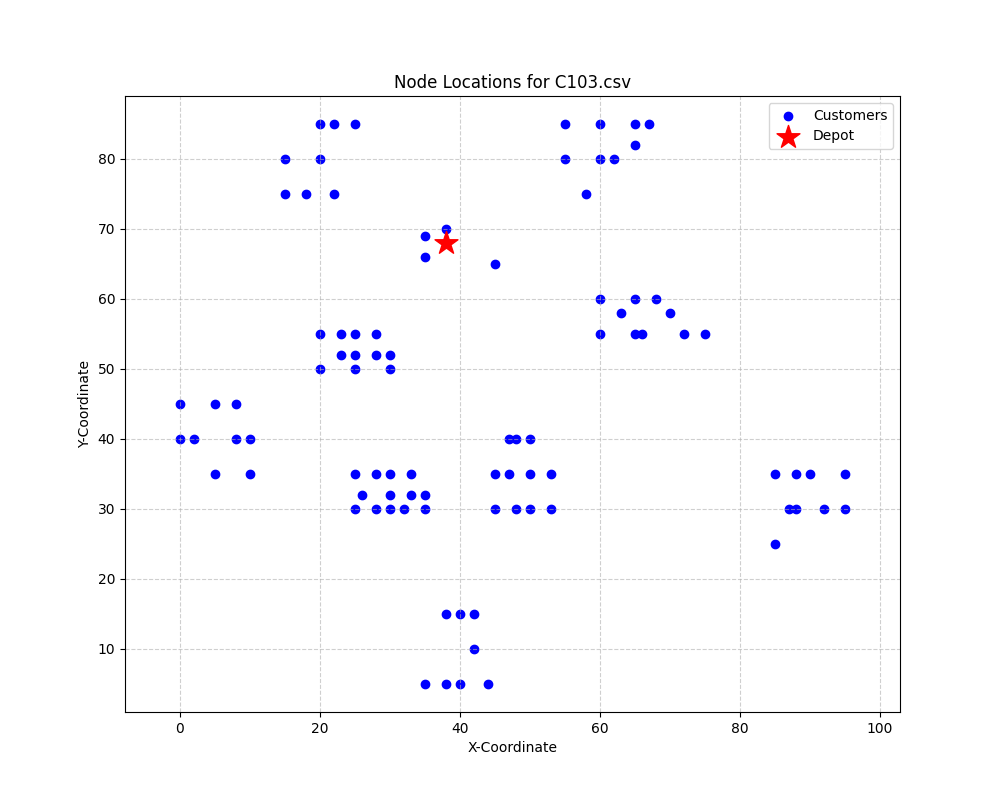}%
  \label{fig:c103}}

\vspace{0.4em}

\subfloat[R101]{%
  \includegraphics[width=0.48\columnwidth]{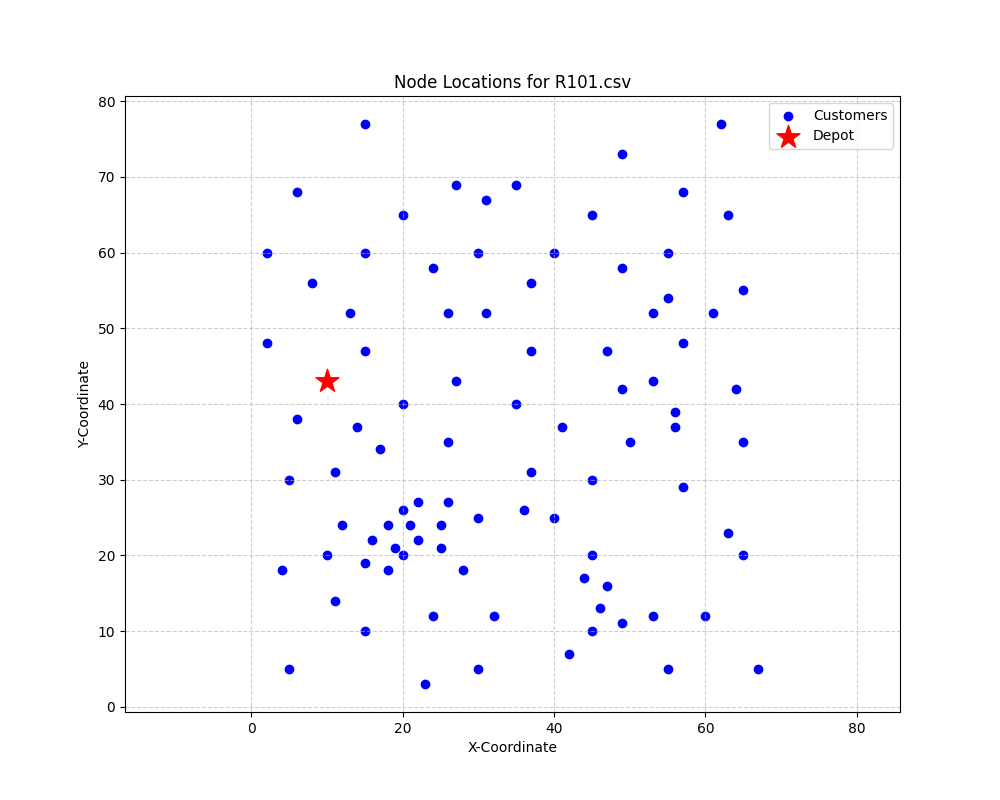}%
  \label{fig:r101}}%
\hfill
\subfloat[R103]{%
  \includegraphics[width=0.48\columnwidth]{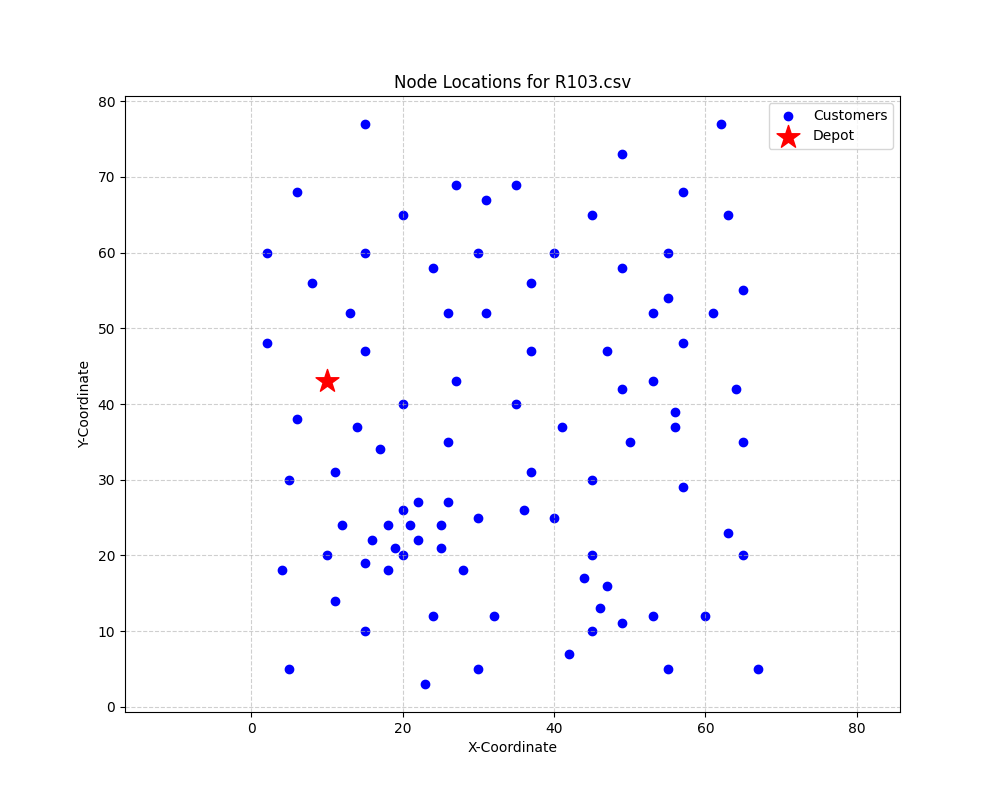}%
  \label{fig:r103}}

\vspace{0.4em}

\subfloat[C201]{%
  \includegraphics[width=0.48\columnwidth]{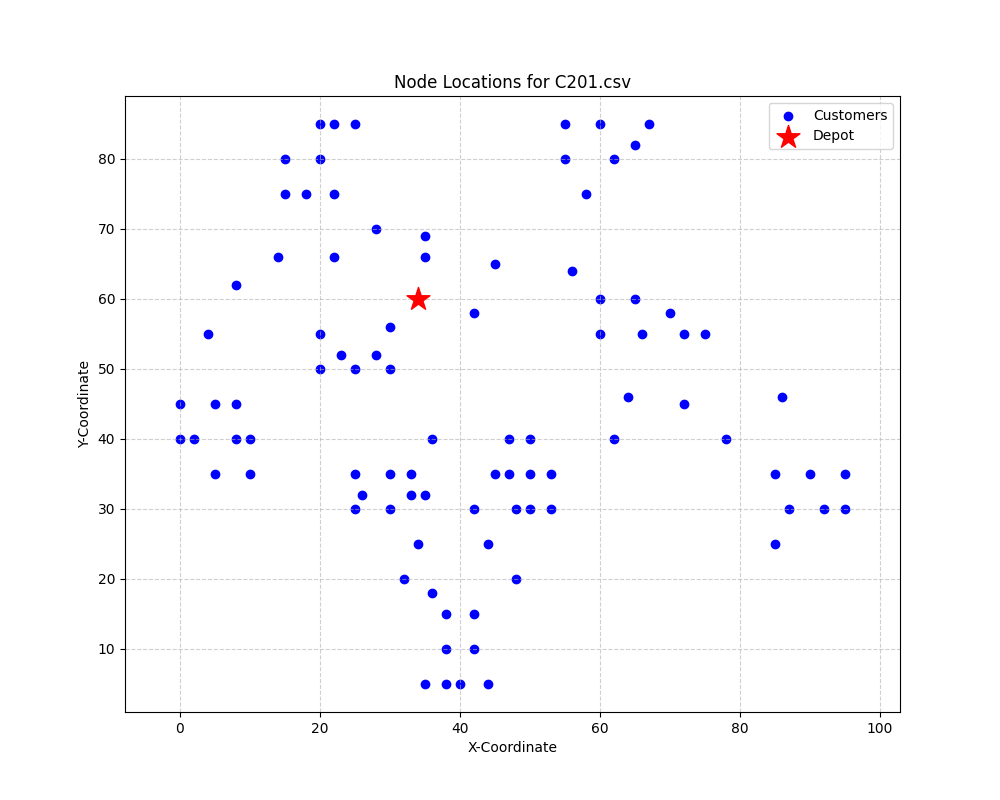}%
  \label{fig:c201}}%
\hfill
\subfloat[C203]{%
  \includegraphics[width=0.48\columnwidth]{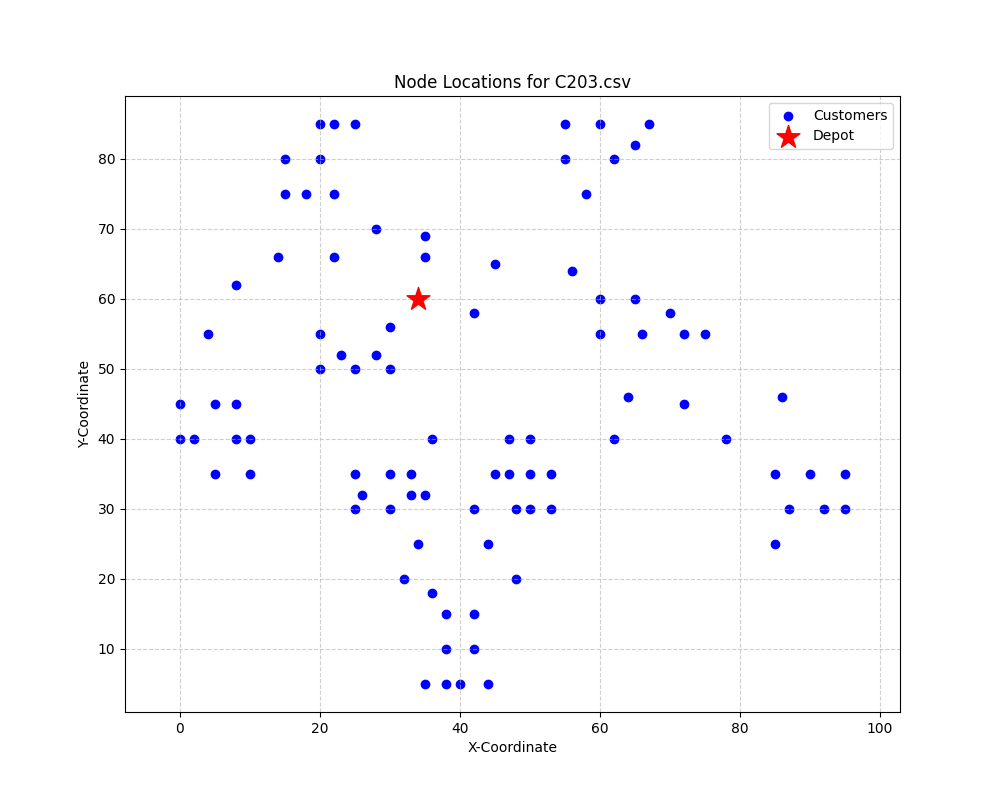}%
  \label{fig:c203}}

\vspace{0.4em}

\subfloat[RC101]{%
  \includegraphics[width=0.48\columnwidth]{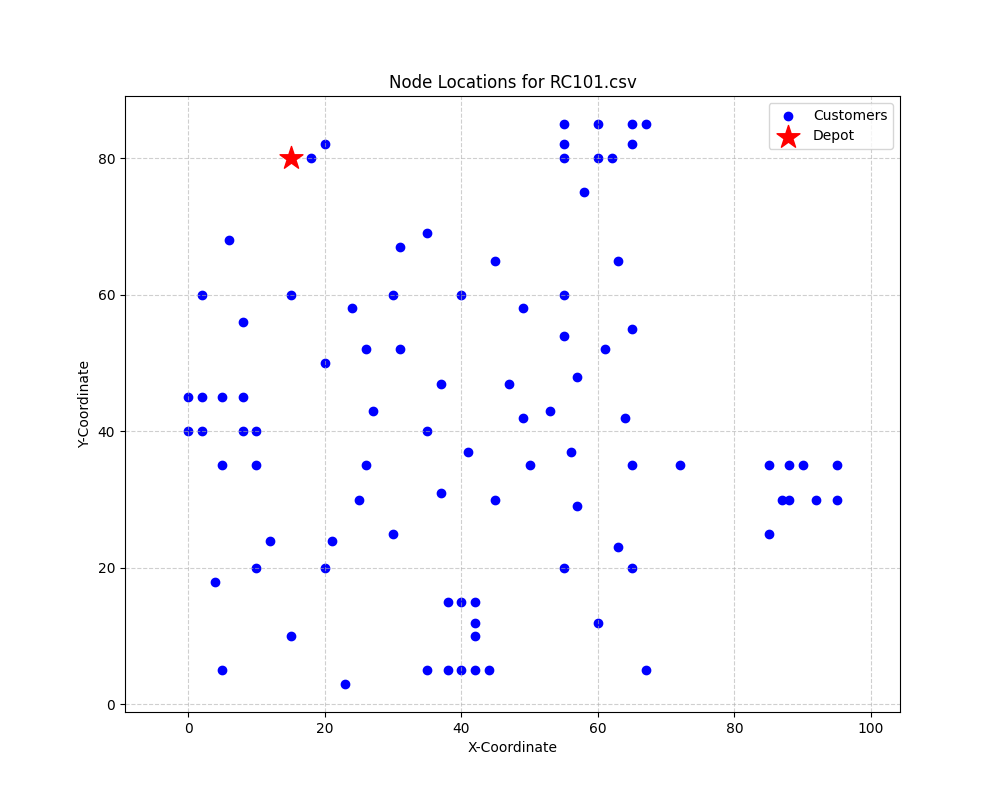}%
  \label{fig:rc101}}%
\hfill
\subfloat[RC103]{%
  \includegraphics[width=0.48\columnwidth]{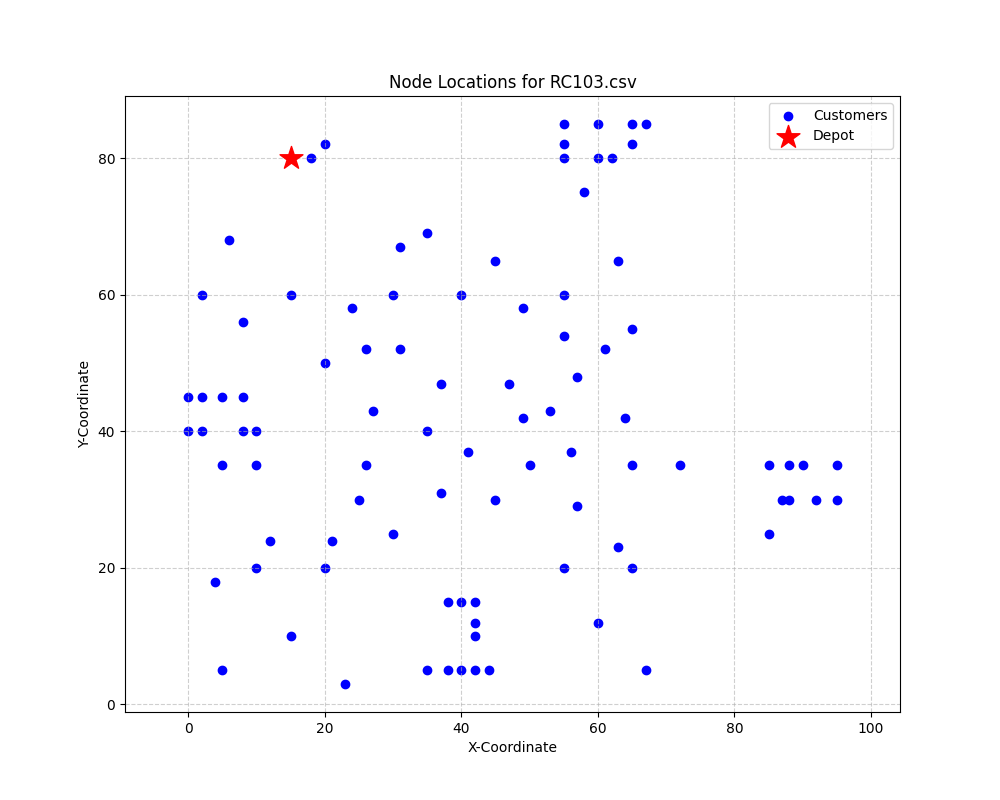}%
  \label{fig:rc103}}

\caption{A subset of instances of the Solomon~\cite{solomon1987} dataset used for the experiments. For clarity, edges are omitted.}
\label{fig:8plots}
\end{figure}

\subsubsection{Problem Instances}
To ensure a fair and standardized evaluation, we used the well-known Solomon VRPTW benchmark datasets. These datasets \cite{solomon1987} are widely used in the academic literature \cite{electronics14040647} \cite{su151410982} \cite{Poohoi2024StasKmeanVRPTW} \cite{YaoMo2025TDVRPSubwayTRISTAN} and are divided into three main categories based on the distribution of customer locations:

\begin{itemize}
  \item \textbf{Type C (Clustered):} Customers are located in distinct, tight clusters.
  \item \textbf{Type R (Random):} Customer locations are uniformly distributed across the service area.
  \item \textbf{Type RC (Random and Clustered):} A hybrid of both, with a mix of randomly distributed and clustered customers.
\end{itemize}

For our experiments, while our evaluation of classical heuristics operates on the standard 100-customer Solomon benchmarks to demonstrate performance at scale, assessing Quantum and Quantum-Inspired solvers requires a distinct approach due to the current limits of combinatorial optimization on these architectures.

\textbf{1.\ Stratified Instance Selection}

To accommodate the varying capacities of our solvers, we adopted a stratified testing strategy:

\begin{itemize}
    \item \textbf{Classical Tier (Greedy and Savings):} These solvers were benchmarked against the full 100-customer instances from the Solomon dataset (C, R, and RC types). This establishes a baseline for algorithm performance on industry-standard, large-scale problems.

    \item \textbf{Quantum Tier (FQS, APS):} Due to the combinatorial explosion inherent in mapping VRP to a QUBO model, solving 100-customer instances directly is currently intractable. For these solvers, we generated iterative sub-instances from the standard Solomon datasets.
\end{itemize}

\textbf{2.\ Sub-Instance Generation}

To ensure reproducibility and realistic data distributions, we did not generate random graphs. Instead, we extracted the first $N$ customers from the standard Solomon CSV files, where $N$ represents the computationally maximum possible threshold of the solver (Scalability Extension). 

\begin{itemize}
    \item This method preserves the spatial clustering and time-window characteristics of the original benchmark families (e.g., Clustered vs.\ Random).
    \item It allows us to precisely identify the ``breaking point'' where the uncoarsened solver fails due to variable explosion, and subsequently demonstrate how graph coarsening restores solution feasibility.
\end{itemize}

\textbf{3.\ Hypothesis Testing Strategy}

 We define the Maximum Solvable Size $N_{\max}$ as the largest number of customers a specific solver configuration can handle within a fixed runtime and memory limit.

\begin{enumerate}
    \item We first determine $N_{\max}$ for the uncoarsened (original) solver formulation.
    \item We then apply our graph coarsening pipeline to instances of size $N > N_{\max}$  to demonstrate that the reduced graph size via graph-coarsening brings the problem back within the solver's effective range.
\end{enumerate}

This methodology allows us to validate the coarsening algorithm's utility across the spectrum: improving speed and solution quality for classical solvers on large graphs, and enabling feasible solutions for quantum solvers on graphs that would otherwise be too complex to map.

\subsubsection{Performance Metrics}
All metrics are computed on the final, inflated routes evaluated against the original uncoarsened graph:
\begin{itemize}
  \item \textbf{Total Distance:} The sum of all travel distances for all vehicles in the solution.
  \item \textbf{Number of Vehicles:} The total number of vehicles required to service all customers.
  \item \textbf{Total Route Duration:} A spatio-temporal metric representing the total time vehicles spend on their routes from depot departure to return. It comprises three components per customer stop:
\begin{itemize}
\item \textit{Travel Time:} Time spent driving from the previous stop, proportional to distance $\tau$.
\item \textit{Waiting Time:} Time a vehicle waits if it arrives before a customer's ready time $e$.
\item \textit{Service Time:} Time spent at the customer location, $s$.
\end{itemize}
  \item \textbf{Feasibility:} A boolean flag indicating that the solution is free of capacity and time-window violations.
  \item \textbf{Validity:} A boolean flag specific to the quantum tier, indicating that the total demand served by a quantum solution equals the total demand served by the OR-Tools reference solution on the same sub-instance. Validity confirms that the QUBO encoding covers all customers correctly, independent of route quality. See Section~\ref{sec:ortools_ref} for the construction of this reference.
  \item \textbf{Scalability Extension:} The ability to solve instances with $N > N_{\max}$.
\end{itemize}

For the hyperparameter tuning process, we defined a single, aggregate Objective Score to guide the search for optimal parameters. This score, defined in Equation~\ref{eq:objective_score}, combines the primary metrics with heavy penalties for any infeasibilities.

\begin{figure}[h]
\centering
\includegraphics[width=0.4\textwidth]{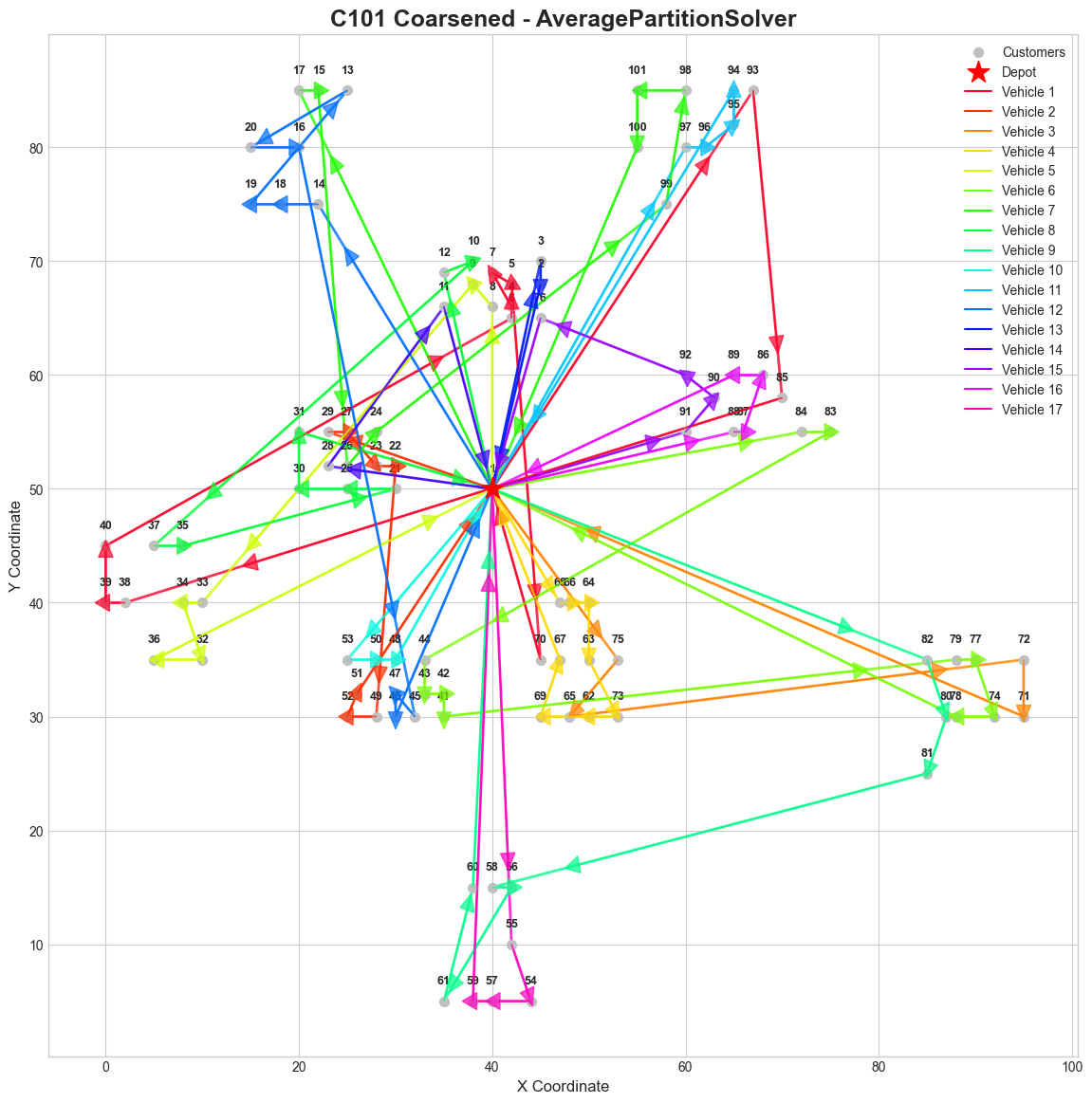}
\caption{Example 100-customer Solomon instance solved using the Average Partitioning Solver with the proposed coarsening procedure.}
\label{fig:example_aps}
\end{figure}

\begin{figure}[h]
\centering
\includegraphics[width=0.4\textwidth]{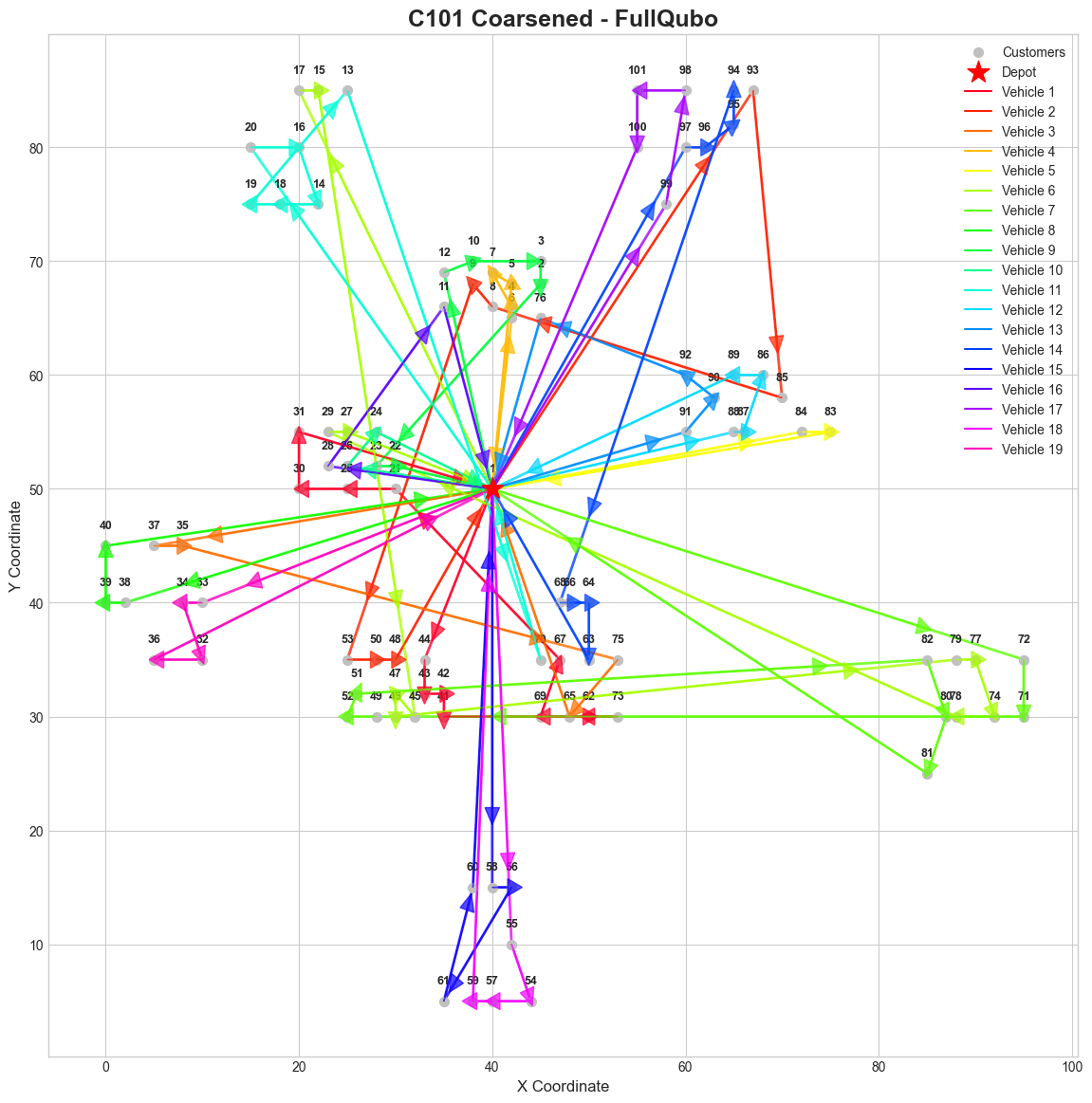}
\caption{Example 100-customer Solomon instance solved using the Full QUBO Solver with the proposed coarsening procedure.}
\label{fig:example_fqs}
\end{figure}

\begin{equation}
\label{eq:objective_score}
Score = \tau_{total} + \lambda_{v} \cdot N_{vehicles} + \lambda_{c} \cdot V_{capacity} + \lambda_{t} \cdot V_{time}
\end{equation}

Here, $\tau_{total}$ is the total distance, $N_{vehicles}$ is the number of vehicles, $V_{capacity}$ and $V_{time}$ are boolean flags (1 for violation, 0 otherwise) for capacity and time window violations, and $\lambda_{v}, \lambda_{c}, \lambda_{t}$ are large penalty weights to ensure that feasible solutions with fewer vehicles are prioritized. For these experiments, we set $\lambda_{v}=1000$, $\lambda_{c}=1000$, and $\lambda_{t}=1000$.

\subsection{Methodology}

\subsubsection{OR-Tools Reference for Sub-Instance Scales}
\label{sec:ortools_ref}

The Solomon benchmark dataset provides best-known solutions for the full 100-customer instances, which are well established in the literature~\cite{solomon1987}. However, no reference solutions exist for the sub-instance scales used in our quantum experiments ($N=5$ and $N=10$ customers), since these truncated instances are specific to our experimental setup and are not part of any standard benchmark set.

To fill this gap, we implemented an OR-Tools CP-SAT solver~\cite{ortools} and ran it on the identical $N=5$ and $N=10$ sub-instances used for the quantum tier, under a 10-second time limit. At these small scales the CP-SAT solver reliably finds provably optimal or near-optimal solutions within the time budget. The resulting OR-Tools solutions serve two roles: (i) they define the \emph{validity} reference: a quantum solution is valid if it serves the same total demand as the OR-Tools solution on the same instance, confirming that the QUBO encoding covers all customers; and (ii) they provide a quality context for interpreting quantum results, without constituting a direct solver competition given the fundamental differences in problem representation and architecture.

\subsubsection{Experimental Phases}

For every instance, we executed a three-phase comparison:

\textbf{1.\ Solver Configurations and Tuning}

We performed a random hyperparameter search (40 trials per instance) across the parameter space:
\[
\begin{aligned}
\alpha  &\in \{0.3, 0.5, 0.7, 0.9, 1.0\} \\
\beta   &\in \{0.2, 0.4, 0.6, 0.8, 1.0\} \\
P       &\in \{0.3, 0.4, 0.5, 0.6, 0.7, 0.8\} \\
R_{\text{coeff}} &\in \{0.5, 1.0, 1.5, 2.0, 3.0, 5.0\}.
\end{aligned}
\]

A key finding from this search is that no single hyperparameter set performs optimally across all instance families. The structural differences between clustered (C-type) and random (R-type) instances; in particular the spatial spread and time-window width analysed in Section~\ref{sec:structure_analysis} require distinct coarsening configurations to achieve best feasibility and solution quality. We therefore adopt a \emph{per-family} hyperparameter strategy: the best-found configuration from the random search is applied separately for each instance family in the main experiments. The selected configurations are reported in Table~\ref{tab:hyperparam_families}.

\begin{table}[htbp]
\centering
\caption{Best coarsening hyperparameters per Solomon instance family, selected by lowest objective score (Eq.~\ref{eq:objective_score}) over the random search.}
\label{tab:hyperparam_families}
\small
\begin{tabular}{lcccc}
\toprule
Family & $P$ & $R_{\text{coeff}}$ & $\alpha$ & $\beta$ \\
\midrule
C-type (C1, C2)   & 0.5 & 2.0 & 1.0 & 1.0 \\
R-type (R1, R2)   & 0.4 & 0.5 & 1.0 & 0.6 \\
\bottomrule
\end{tabular}
\end{table}

\textbf{2.\ Uncoarsened Baseline}

Then we executed the solver (Classical or Quantum) on the original graph. For the Quantum track, this step established the Maximum Solvable Size ($N_{\max}$): The largest instance size the solver could handle before failing due to memory constraints.

\textbf{3.\ Coarsened Pipeline}

We last applied our Spatio-Temporal Coarsening algorithm to reduce the graph size before feeding it to the solver. The resulting routes were subsequently inflated and evaluated.

We utilized the FQS and APS formulations running on a classical simulated annealing backend. Unlike the classical algorithms, the quantum algorithms required targeted penalty tuning rather than random search. To enforce hard constraints on the uncoarsened graphs, we established a strict penalty hierarchy:

\begin{equation}
  \begin{aligned}
    P_{\text{unique}}     &= 10\lambda = 10{,}000{,}000 \\
    P_{\text{self-loop}}  &= 5\lambda  =  5{,}000{,}000 \\
    P_{\text{cap}}        &= 5\lambda  =  5{,}000{,}000 \\
    P_{\text{tw}}         &= 3\lambda  =  3{,}000{,}000 \\
    P_{\text{continuity}} &= \lambda   =  1{,}000{,}000 \\
    P_{\text{tw,soft}}    &= 0.15\lambda = 150{,}000   \\
    P_{\text{veh-start}}  &= 0.1\lambda =  100{,}000   \\
    w_{\text{dist}}       &= 10^{-4}\lambda = 100
  \end{aligned}
  \label{eq:penalty_hierarchy}
\end{equation}

The dominant term $P_{\text{unique}}$ enforces that every customer is visited
exactly once; its value is set to twice the capacity and self-loop penalties to
ensure that visit-uniqueness violations are always more costly than any other
infeasibility. $P_{\text{self-loop}}$ and $P_{\text{continuity}}$ are derived
internally as $0.5\,P_{\text{unique}}$ and $0.1\,P_{\text{unique}}$ respectively,
encoding route-structural constraints. The time-window penalty $P_{\text{tw}}$
penalises provably infeasible consecutive-visit pairs identified during
pre-processing; a secondary soft penalty $P_{\text{tw,soft}} = 0.05\,P_{\text{tw}}$
is applied to temporally tight (though not strictly infeasible) transitions. The
vehicle start cost $P_{\text{veh-start}}$ discourages unnecessary route creation,
while the distance weight $w_{\text{dist}}$ preserves the routing objective within
the energy landscape without competing with the hard constraint terms.
This hierarchy was critical for coercing the solvers into feasible regions
within the complex energy landscape of larger sub-instances.

\begin{figure}[htbp]
\centering
\includegraphics[width=0.9\linewidth,height=0.30\textheight]{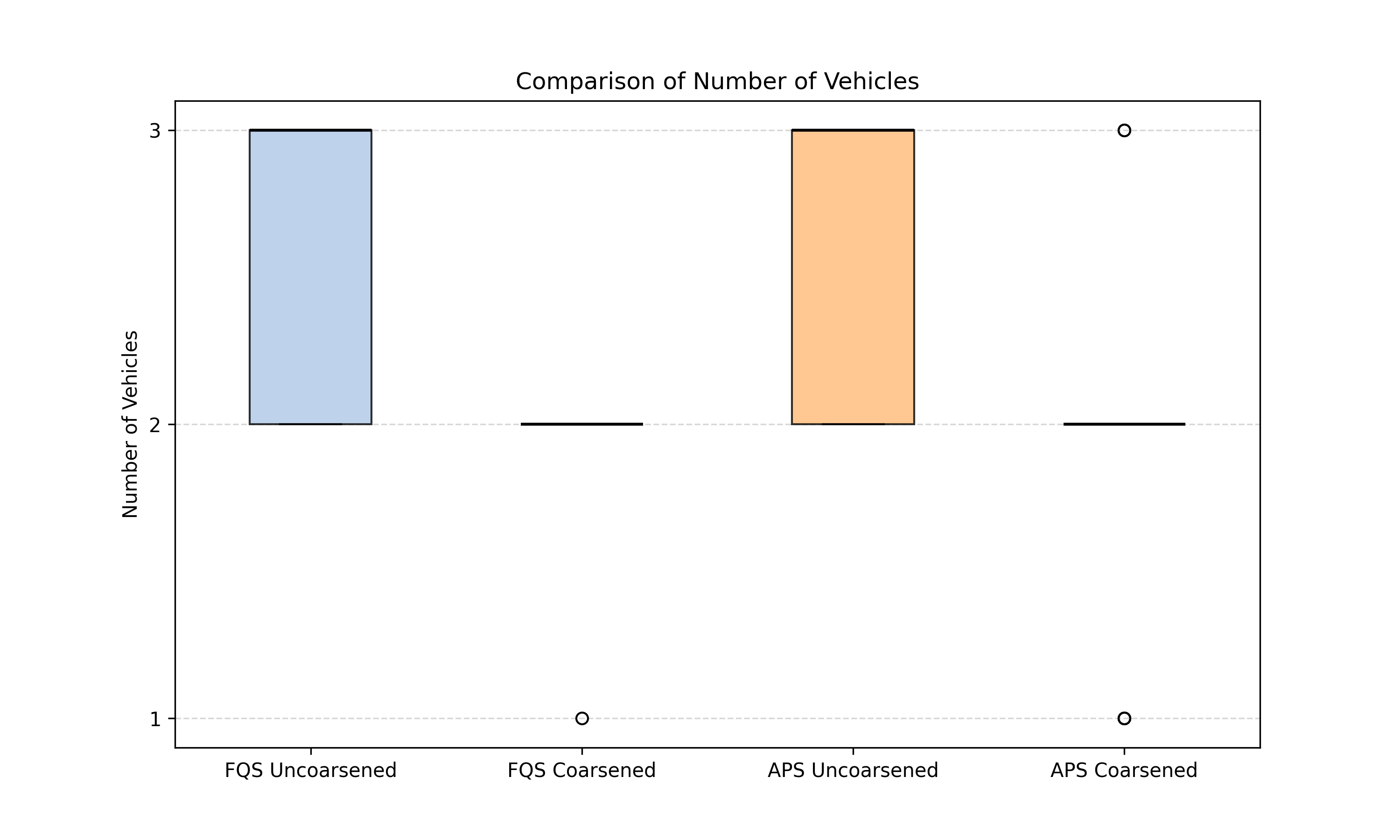}
\caption{Comparison of the number of vehicles required across uncoarsened and inflated solutions for both FQS and APS solvers. Uncoarsened solutions show a larger spread and generally higher vehicle counts due to local infeasibilities. After inflation, the number of vehicles decreases significantly, indicating that the coarsening process successfully aggregates demand and leverages the reduced search space to find more efficient routes that better utilize vehicle capacity.}
\label{fig:box_num_vehicles}
\end{figure}

\begin{table}[htbp]
\centering
\caption{Selected Subset of the Solver Improvements}
\label{tab:selected_subset}

\small
\setlength{\tabcolsep}{3pt}

\begin{tabular}{lcccccc}
\toprule
Inst. & Solver & Dist. & Dur. & Veh. & C/F & CT \\
\midrule
C108 & Greedy   & 36.57\% & 34.37\% & 31.25\%  & 0/23\%   & 87.14\% \\
C108 & Savings  & -49.91\% & -34.51\% & -54.55\% & 0/0\%    & 82.26\% \\

\midrule
C207 & AvgPart  & 41.77\% & 43.18\% & 20.00\%  & F$\to$T   & 92.86\% \\
C207 & FullQubo & 40.36\% & 51.46\% & 40.00\%  & F$\to$F   & 92.86\% \\
C208 & AvgPart  & 41.65\% & 20.73\% & 0.00\%   & F$\to$T   & 95.16\% \\
C208 & FullQubo & 40.18\% & 41.83\% & 50.00\%  & F$\to$T   & 95.27\% \\
\bottomrule
\end{tabular}

\textbf{Note:} Dist.\ = percentage distance improvement; Dur.\ = percentage route duration improvement; Veh.\ = reduction in the number of vehicles used; C/F = auxiliary feasibility or constraint indicator reported for the corresponding solver; CT = percentage reduction in computation time. Positive values indicate improvements, while negative values indicate deterioration compared to the uncoarsened solution.
\end{table}

\begin{figure}[htbp]
\centering
\includegraphics[width=0.9\linewidth,height=0.30\textheight]{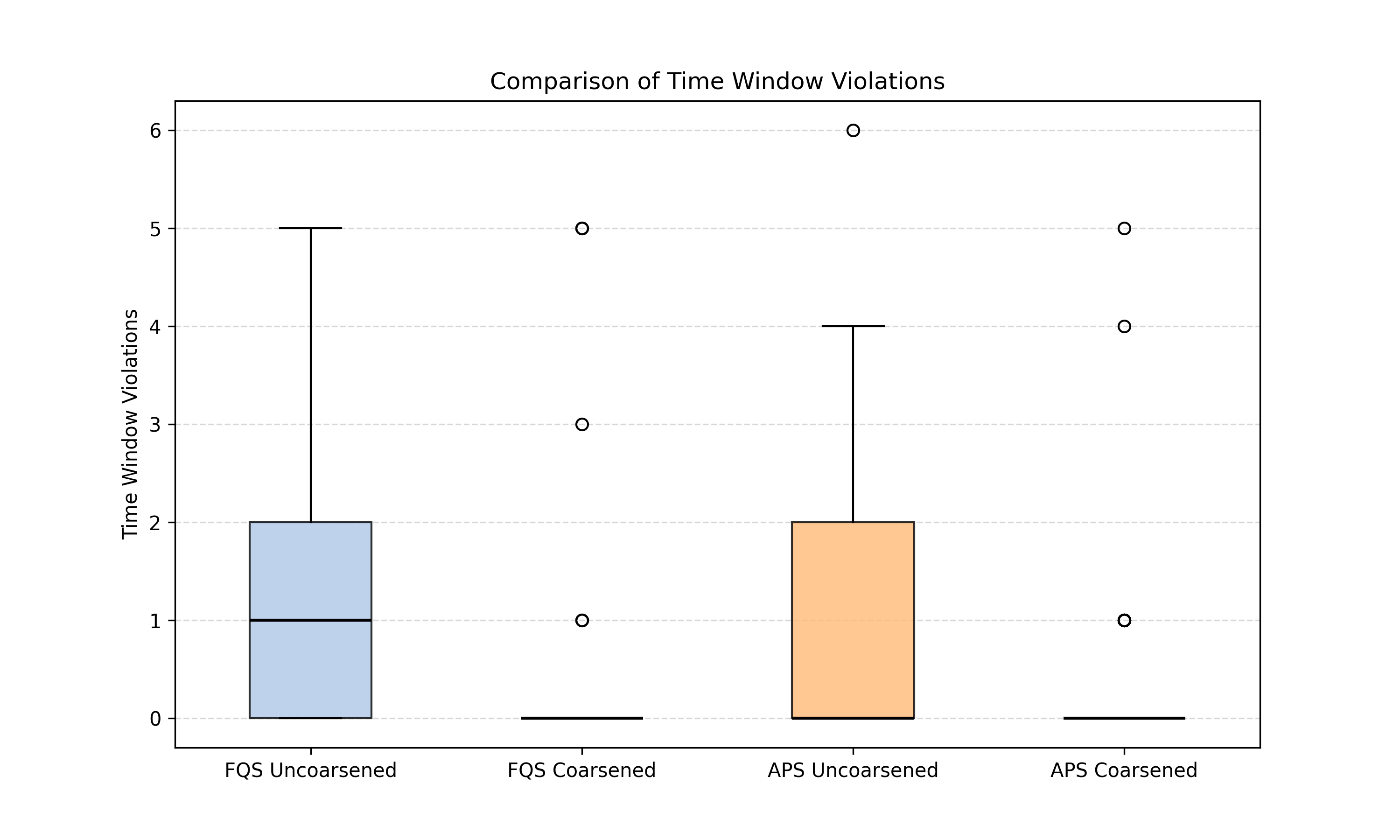}
\caption{Distribution of time window violations for uncoarsened and inflated solutions under Full Qubo Solver and Average Partitioning Solver. Coarsened runs exhibit lower violation counts.}
\label{fig:box_time_window}
\end{figure}

\begin{figure}[htbp]
\centering
\includegraphics[width=0.9\linewidth,height=0.30\textheight]{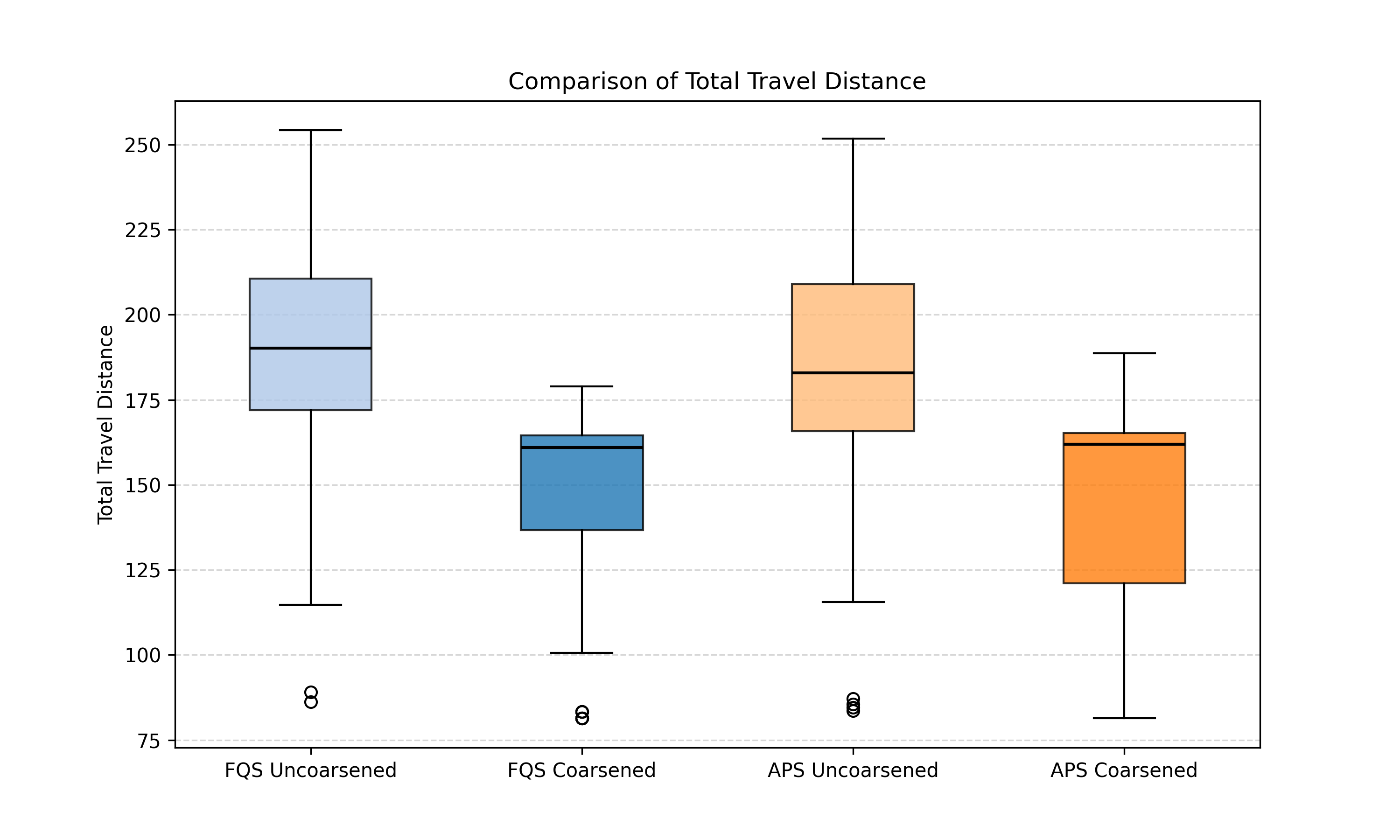}
\caption{Comparison of total travel distance for FQS and APS before and after inflation. While uncoarsened solutions show higher variability, inflated routes achieve lower and more consistent total distances. This suggests that the proposed spatio--temporal coarsening preserves near-optimal spatial structure, which the inflation step further refines for improved route efficiency.}
\label{fig:box_total_distance}
\end{figure}

\begin{figure}[htbp]
\centering
\includegraphics[width=0.9\linewidth,height=0.30\textheight]{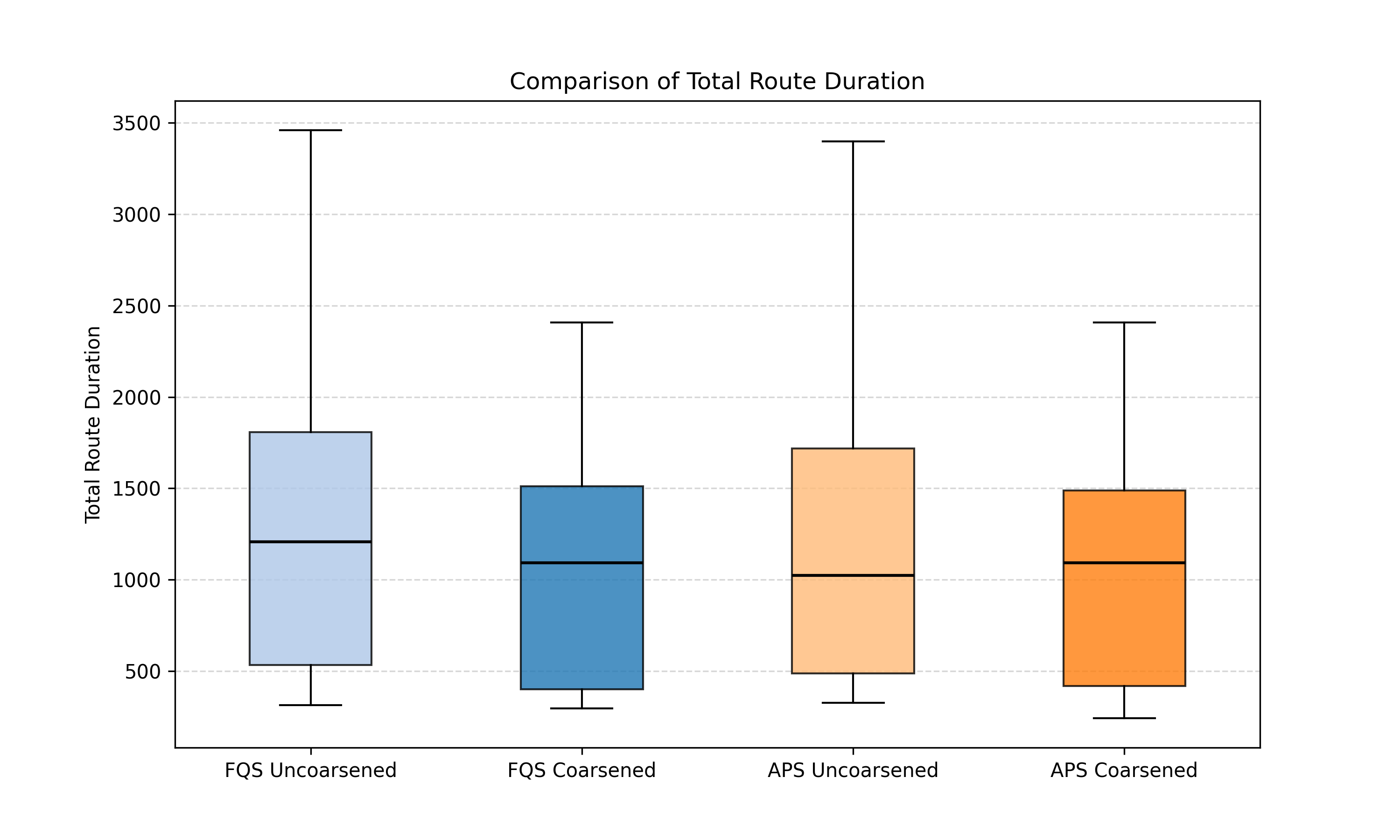}
\caption{Total route duration comparison for uncoarsened and inflated results across both heuristics. Route duration includes travel, waiting, and service times. Mean numbers being almost the same as their uncoarsened counterparts, meaning loss of solution quality after coarsening operation is minimal.}
\label{fig:box_route_duration}
\end{figure}

\subsection{Results and Discussion}

The experiments are organised into two intentionally separate tiers, each addressing a distinct research question. The classical tier asks: \textit{does coarsening improve solution quality and speed for established heuristics?} The quantum tier asks: \textit{does coarsening extend the feasible operating range of QUBO-based solvers by reducing their runtime?} The common thread is the coarsening pipeline. In both tiers the baseline is the same solver operating on the uncoarsened graph. The two tiers are not intended as a direct solver-to-solver comparison; doing so would be methodologically unsound given the fundamental differences in problem scale and solver architecture.

Our goal is therefore to investigate whether \textit{graph coarsening} improves each solver relative to its own uncoarsened baseline. The coarsening procedure serves as a structural preconditioner that helps the solver navigate a reduced search space more effectively, producing higher-quality routes in less time. This comparison isolates the contribution of the coarsening strategy itself, independent of the solver’s inherent limitations.

\subsubsection{Overall Performance}
\begin{figure}[!t]
    \centering
    \includegraphics[width=3.3in]{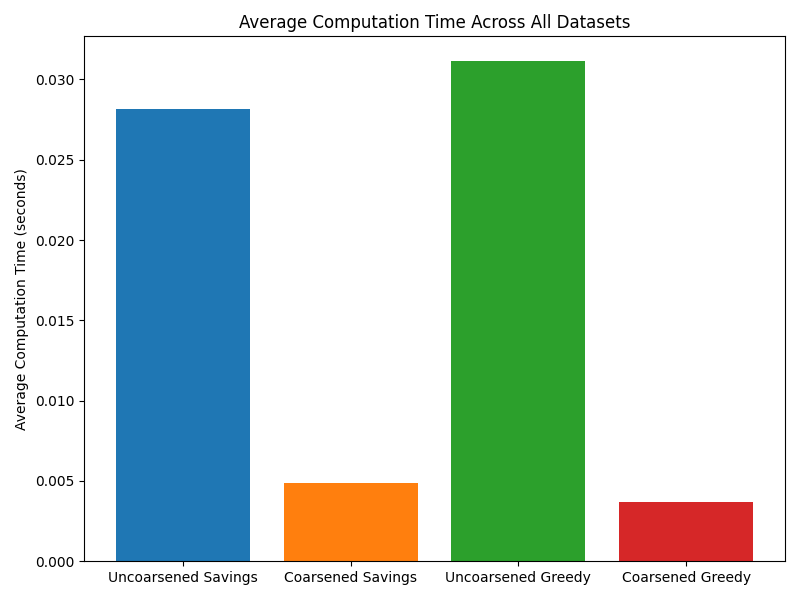}
    \caption{Average computation time comparison between Uncoarsened and Coarsened methods
    for both Savings and Greedy solutions. Graph coarsening consistently reduces computation time.}
    \label{fig:computation-time}
\end{figure}

The primary finding from our experiments is that the effect of graph coarsening on solution quality is strongly dependent on the solver used. For the Greedy heuristic, coarsening consistently improves all measured metrics. Across all 45 tested instances (C-, R-, and RC-type), coarsening improved total distance in 41 cases (91.1\%), with a median improvement of 36.57\%. The effect is robust across instance families: the median distance improvement is 31.72\% for clustered (C-type), 29.26\% for random (R-type), and 45.39\% for mixed (RC-type) instances. The median reduction in the number of vehicles deployed is 33.33\%, and computation time is reduced by a median of 87.34\% (Figure~\ref{fig:computation-time}). In no instance did coarsening introduce capacity or time-window violations for the Greedy heuristic. A Wilcoxon signed-rank test on the 56 paired distance observations confirms that this improvement is statistically significant ($W = 33$, $p < 0.001$), with a consistent positive effect across all instance families.

The Savings heuristic presents a starkly different picture: coarsening degrades solution distance in all 45 tested instances (median $-33.17\%$), because aggregation distorts the pairwise savings scores on which the algorithm depends, and introduces new time-window violations in several instances where the uncoarsened solution was previously violation-free. Computation time is nonetheless reduced by a median of 82.44\%, making coarsening viable for the Savings algorithm when throughput matters more than solution quality. A Wilcoxon signed-rank test on the 56 paired distance observations confirms that the degradation is also statistically significant ($W = 0$, $p < 0.001$), underscoring the structural sensitivity of the Savings heuristic to graph aggregation.

The divergence between the two classical solvers might be explained by their fundamentally different search strategies. The Greedy heuristic builds routes locally, customer by customer, and benefits from the reduced search space imposed by coarsening. The Savings heuristic performs a global merge optimization that is sensitive to the exact pairwise distance structure of the graph; forced aggregation of customers into super-nodes distorts this structure and degrades the quality of the computed savings scores, leading to suboptimal merge decisions that persist after inflation.
TO BE PROVEN

\subsubsection{Quantum Tier: Validity and Feasibility}
\label{sec:quantum_results}

Experiments were conducted on all 56 Solomon instances at $N=5$ and $N=10$ customers using four solver configurations: Uncoarsened FQS, Uncoarsened APS, Inflated (coarsened) FQS, and Inflated APS  yielding 224 solver-instance pairs per scale. All Inflated results use the per-family hyperparameters reported in Table~\ref{tab:hyperparam_families}; Uncoarsened configurations are family-agnostic by construction.

\paragraph{Validity.}
Across all 448 solver-instance pairs, every quantum solution achieves 100\% validity: the total demand served matches the OR-Tools reference on every instance, at both scales and under all four configurations. This confirms that the QUBO formulation correctly covers all customers in every run.

\paragraph{Feasibility at $N=5$.}
At the five-customer scale, all four configurations are 100\% feasible across all 56 instances. Inflated variants average $3.3\,\text{s}$ computation time, compared with $7.1\,\text{s}$ for Uncoarsened a 54\% reduction attributable solely to the coarsening pipeline. Both figures lie below the OR-Tools 10-second time budget, confirming that at this scale all approaches are practically competitive in runtime.

\paragraph{Feasibility at $N=10$.}
Uncoarsened FQS and APS remain near-fully feasible (100\% and 98\% respectively, with one infeasible APS case on RC105). Inflated variants show degraded feasibility: Inflated FQS achieves 75\% feasibility (14 infeasible out of 56) and Inflated APS 80\% (11 infeasible). Infeasibility is strongly family-dependent:

\begin{itemize}
  \item \textbf{C-type (clustered):} Both Inflated FQS and Inflated APS are 100\% feasible across all 17 instances.
  \item \textbf{R-type (random):} Using the R-type hyperparameter set ($P=0.4$, $R_{\text{coeff}}=0.5$, $\alpha=1.0$, $\beta=0.6$), 20 of 23 instances are feasible under Inflated FQS (87\%). The earlier global-hyperparameter run produced only 57\% feasibility on R-type, demonstrating the importance of per-family tuning.
\end{itemize}

The structural reason for residual R-type infeasibility and the sensitivity to hyperparameter choice is analysed in Section~\ref{sec:structure_analysis}.

\paragraph{Coarsening effect on C-type instances at $N=10$.}
For clustered instances, coarsening delivers consistent simultaneous improvements across all three primary metrics (Table~\ref{tab:coarsening_effect_c}). Inflated FQS reduces the average vehicle count by $1.00$ relative to Uncoarsened FQS ($3.00 \to 2.00$), route duration by $27.1\%$, and computation time by $87.1\%$. Inflated APS reduces vehicles by $0.88$, duration by $22.9\%$, and computation time by $87.2\%$. This triple benefit with zero feasibility loss makes C-type instances the demonstrably strongest use case for the coarsening pipeline.

\begin{table}[htbp]
\centering
\caption{Coarsening effect on C-type instances at $N=10$ (Inflated $-$ Uncoarsened, averaged over 17 instances).}
\label{tab:coarsening_effect_c}
\small
\begin{tabular}{lrrr}
\toprule
Solver & $\Delta$Vehicles & $\Delta$Duration & $\Delta$Comp.\ Time \\
\midrule
Inflated FQS & $-1.00$ & $-27.1\%$ & $-87.1\%$ \\
Inflated APS & $-0.88$ & $-22.9\%$ & $-87.2\%$ \\
\bottomrule
\end{tabular}
\end{table}

\paragraph{Coarsening effect on R-type and RC-type instances at $N=10$.}
With the per-family R-type hyperparameters ($P=0.4$, $R_{\text{coeff}}=0.5$, $\alpha=1.0$, $\beta=0.6$), Inflated FQS achieves 87\% feasibility on R-type (20/23 instances), compared with 57\% under the global hyperparameter set. The smaller radius coefficient deliberately restricts merges to the most compatible node pairs, sacrificing coarsening depth in favour of post-inflation feasibility.

\paragraph{Quality context from OR-Tools reference.}
Across both scales, all quantum configurations use more vehicles and produce longer route durations than the OR-Tools reference (Table~\ref{tab:vs_ortools_ref}). At $N=5$, Inflated FQS averages $1.98$ vehicles versus OR-Tools' $1.04$; at $N=10$, $2.05$ versus $1.45$. The gap on route duration follows the same pattern. These differences reflect the inherent limitations of the QUBO/simulated-annealing architecture at these problem sizes, not a deficiency of the coarsening pipeline: the Inflated results are consistently \emph{better} than the Uncoarsened quantum baseline. The one metric where Inflated quantum solvers are competitive with OR-Tools is computation time: at $3.3$--$3.4\,\text{s}$, Inflated variants are approximately $3\times$ faster than the OR-Tools 10-second budget. Uncoarsened variants at $N=10$ average $23\,\text{s}$, $2.3\times$ \emph{slower} than OR-Tools, underlining that coarsening is necessary for quantum solvers to be runtime-competitive even at small scales.

\begin{table}[htbp]
\centering
\caption{Average metrics: Inflated FQS vs.\ OR-Tools reference across all 56 instances.}
\label{tab:vs_ortools_ref}
\small
\begin{tabular}{lrrrr}
\toprule
 & \multicolumn{2}{c}{$N=5$} & \multicolumn{2}{c}{$N=10$} \\
\cmidrule(lr){2-3}\cmidrule(lr){4-5}
 & Inflated FQS & ORT & Inflated FQS & ORT \\
\midrule
Avg.\ vehicles        & 1.98 & 1.04 & 2.05 & 1.45 \\
Avg.\ route duration  & 859  & 662  & 1364 & 975  \\
Avg.\ comp.\ time (s) & 3.3  & 10.0 & 3.4  & 10.0 \\
Feasibility           & 100\% & 100\% & 75\% & 100\% \\
\bottomrule
\end{tabular}
\end{table}

\begin{figure}[htbp]
\centering
\includegraphics[width=0.9\linewidth]{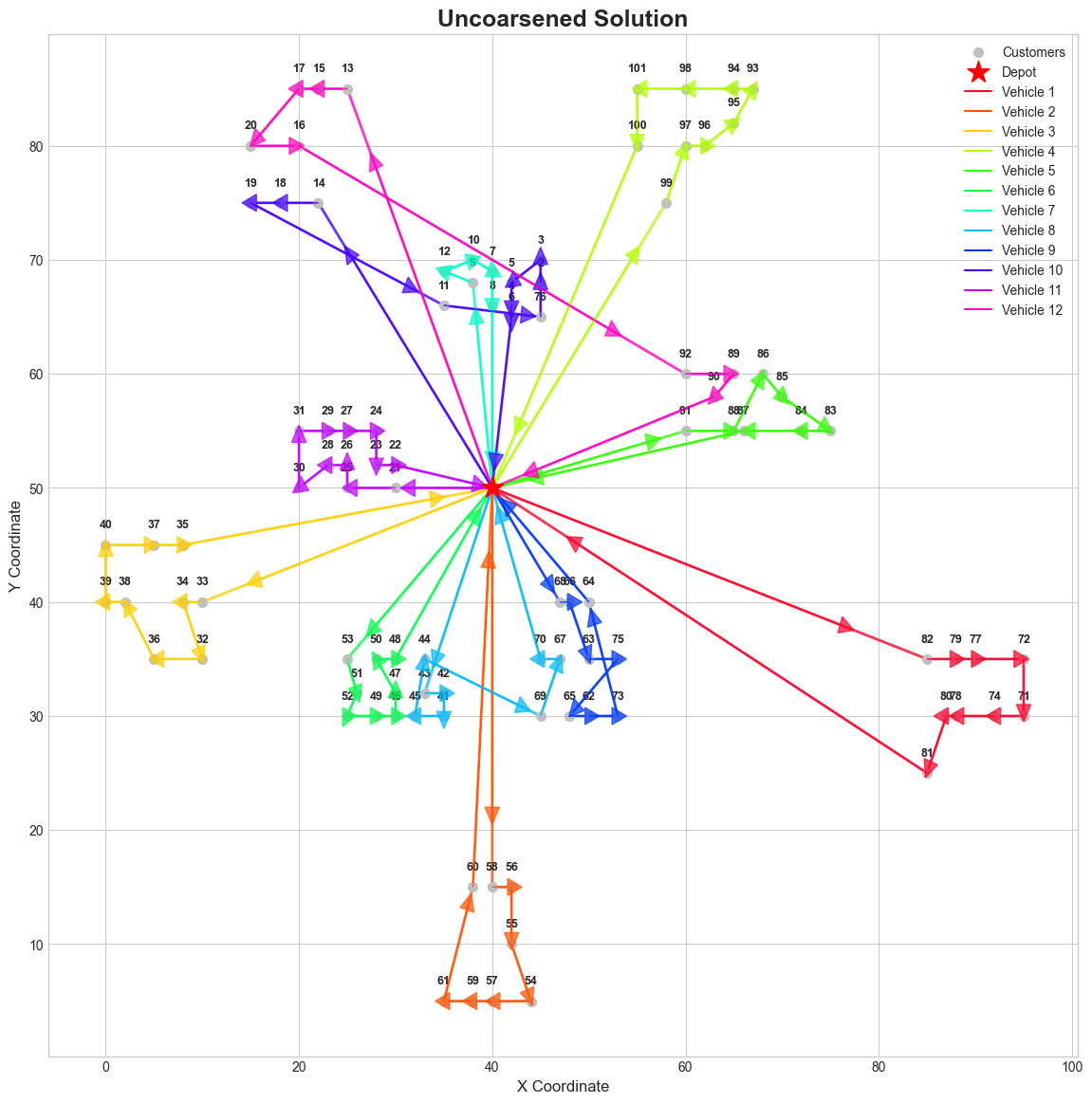}
\caption{Example of an uncoarsened solution.}
\label{fig:unc_sav10}
\end{figure}

\begin{figure}[htbp]
\centering
\includegraphics[width=0.8\linewidth]{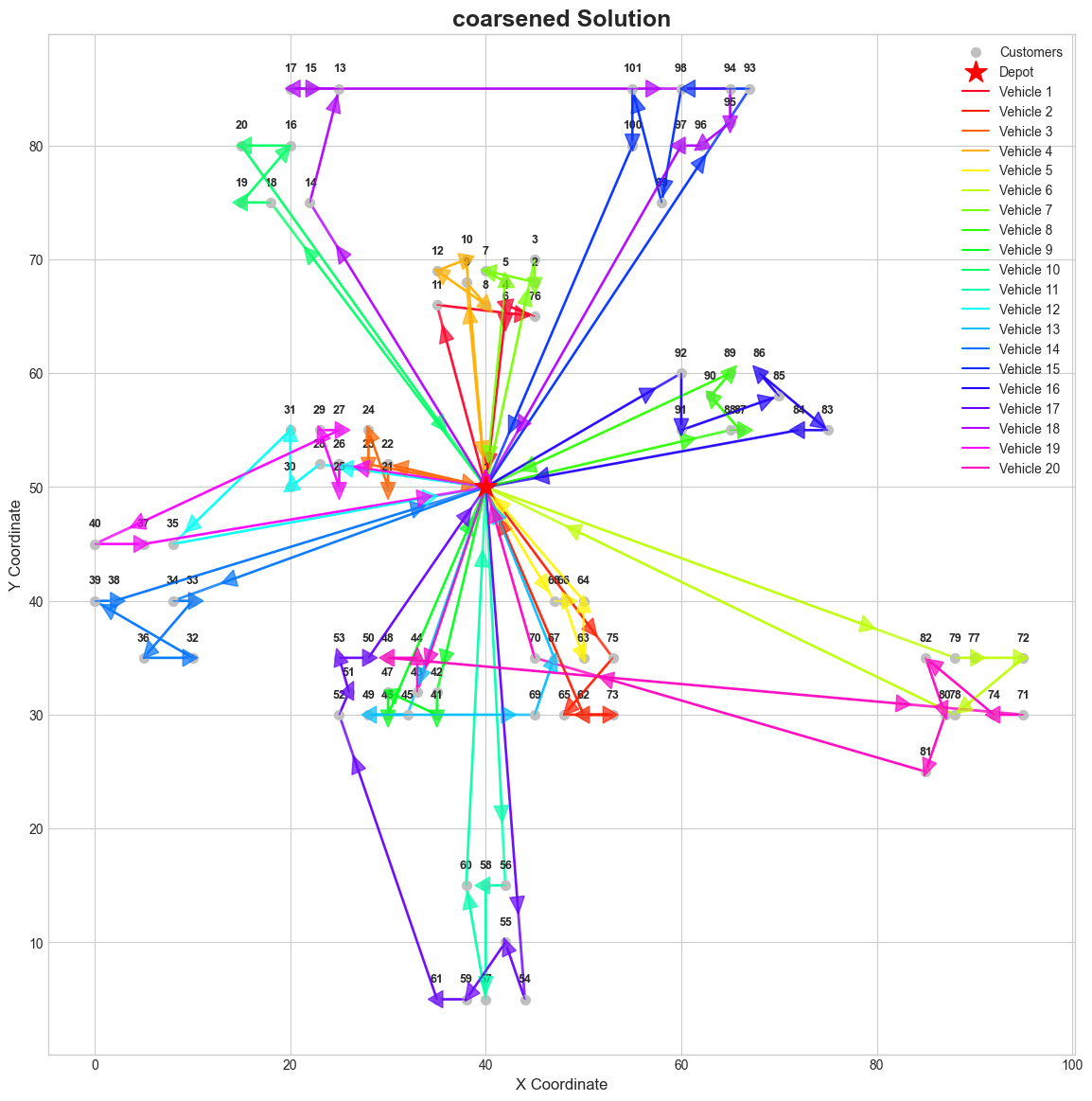}
\caption{Solution after coarsening procedure.}
\label{fig:coars_sav10}
\end{figure}

\begin{figure}[htbp]
\centering
\includegraphics[width=0.7\linewidth]{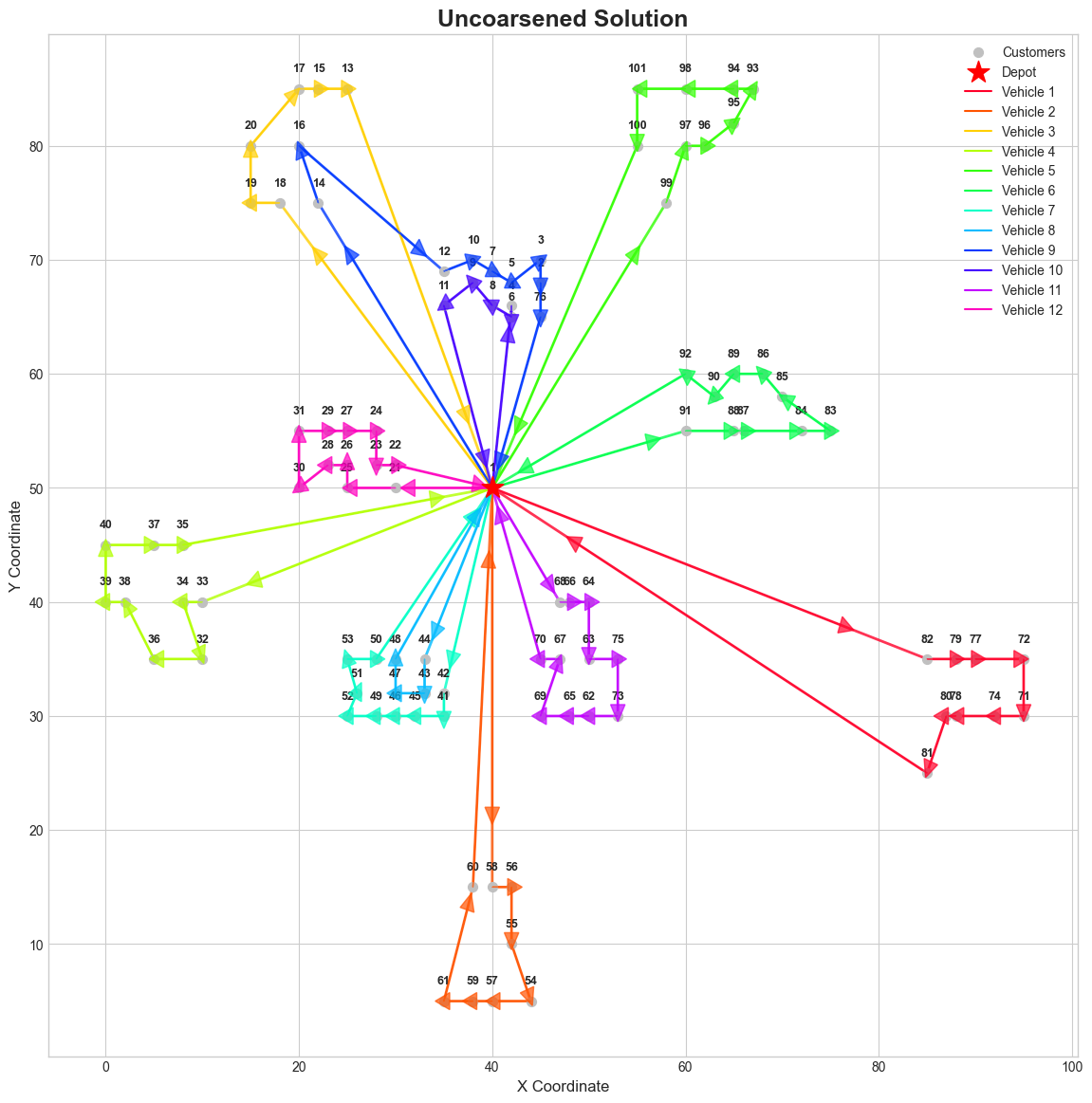}
\caption{Example of an uncoarsened solution.}
\label{fig:unc_sav11}
\end{figure}

\begin{figure}[htbp]
\centering
\includegraphics[width=0.8\linewidth]{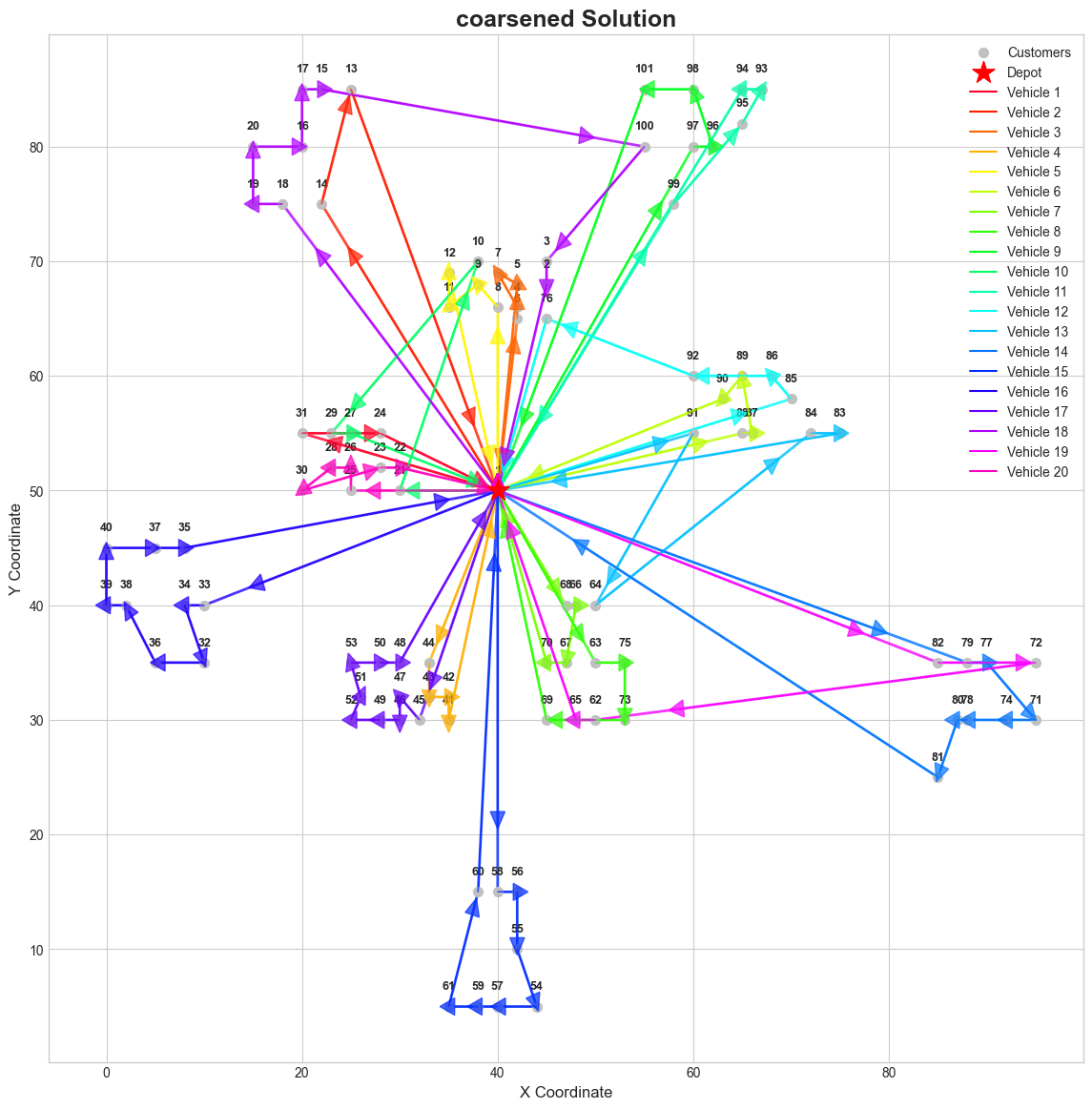}
\caption{Solution after coarsening procedure.}
\label{fig:coars_sav11}
\end{figure}

\subsubsection{Scalability and Feasibility at $N=5$}

At the five-customer scale, all four solver configurations achieve 100\% feasibility and 100\% validity across all 56 instances (224 solver-instance pairs). This confirms that at $N=5$ the QUBO formulation operates well within the solver's capacity, and coarsening's primary contribution at this scale is speed rather than feasibility recovery.

Computation time reductions from coarsening are family-dependent: $-69\%$ on C-type (default hyperparameters) and $-40\%$ on R-type (per-family hyperparameters, Table~\ref{tab:hyperparam_families}). The smaller reduction on R-type reflects that the R-type hyperparameter set uses a smaller radius coefficient ($R_{\text{coeff}}=0.5$), deliberately limiting the number of merges to preserve temporal feasibility; the coarsened graph therefore remains closer in size to the original, trading coarsening depth for solution quality. See Section~\ref{sec:structure_analysis} for the structural explanation.

A broader observation across both solver classes is that computation time reductions from coarsening are larger for quantum solvers than for classical ones. This is a direct consequence of the quadratic scaling of QUBO formulations: reducing the number of customers from $N$ to $N_c = \lfloor P \cdot N \rfloor$ reduces the number of binary decision variables by a factor of approximately $P^2$ and the number of QUBO interaction terms by approximately $P^4$. Coarsening therefore delivers disproportionately larger speedups for QUBO-based solvers operating near their computational limits.

\subsubsection{Scalability Extension: from $N_{\max}$ to $N=100$}
\label{sec:scalability_ext}

Without coarsening, the quantum solvers encounter a combinatorial explosion in the number of QUBO variables as $N$ grows, and the uncoarsened formulation eventually fails to return feasible solutions within the runtime and memory budget.

By applying the spatio-temporal coarsening pipeline to the full 100-customer Solomon instances, the coarsened graph is reduced to a problem size within the solver's effective range. Preliminary results on C-type instances demonstrate that coarsening enables the quantum solvers to find feasible solutions on all 100-customer instances — a scale that is otherwise entirely intractable for these solvers without preprocessing.

This scalability extension is the most practically significant benefit of the coarsening pipeline for quantum architectures: it transforms an otherwise combinatorially intractable problem into a solvable one by leveraging the spatial and temporal structure of the instance to compress the search space, with the inflation step recovering a route over the full original customer set.

Complete per-instance results for all 56 Solomon instances and all solver configurations are available at \url{https://github.com/mkingmking/graph-coarsening}.

\subsubsection{Instance Structure and Coarsening Compatibility}
\label{sec:structure_analysis}

The divergence in coarsening effectiveness across instance families motivates a structural analysis. Table~\ref{tab:structure} summarises the spatial and temporal characteristics of the first $N=10$ customers across all six Solomon families.

\begin{table}[htbp]
\centering
\caption{Spatial and temporal structure of Solomon families (first 10 customers per instance, averaged over instances in each family).}
\label{tab:structure}
\small
\begin{tabular}{lrrrr}
\toprule
Family & Spatial $\sigma$ & Avg.\ TW width & RT $\sigma$ & TW overlap \\
\midrule
C1  &  3.4 &  341 &  285 & 42.5\% \\
C2  & 16.8 &  988 & 1057 & 45.3\% \\
R1  & 21.8 &   93 &   45 & 78.5\% \\
R2  & 21.8 &  477 &  215 & 79.6\% \\
RC1 & 18.1 &   89 &   39 & 84.7\% \\
RC2 & 18.1 &  387 &  197 & 74.2\% \\
\bottomrule
\end{tabular}
\end{table}

C-type instances exhibit tight spatial clustering ($\sigma \approx 3$--$17$) and wide time windows (up to 988 time units). Spatially adjacent customers are therefore also likely to share sufficient temporal overlap to form valid super-nodes. R-type and RC-type instances are spatially dispersed and have narrow windows: R1 median window width is 10 time units — equal to service time, leaving zero temporal slack. The tightest instance (R101) has only 8.9\% pairwise time-window overlap, meaning almost no customer pair can be merged without violating at least one individual window upon inflation. Notably, R-type customers are \emph{more} temporally concentrated than C-type (ready-time $\sigma = 45$ vs.\ $285$), but near-zero window widths negate this proximity: customers cluster around similar service times but have no mutual slack to absorb.

This analysis confirms that coarsening effectiveness is jointly determined by spatial clustering and time-window width, not by either dimension alone. Instances where customers are both spatially close and temporally flexible constitute the natural target domain of the proposed pipeline. Crucially, however, targeted hyperparameter tuning does substantially recover R-type feasibility: by reducing $R_{\text{coeff}}$ from 2.0 to 0.5 and adjusting the temporal weight $\beta$, the coarsener is restricted to merging only the most compatible node pairs, raising R-type feasibility from 57\% to 87\% at $N=10$. The remaining infeasible instances represent cases where no merge configuration can produce a coarsened graph whose solution inflates feasibly, owing to the complete absence of pairwise time-window overlap in those specific sub-instances.

\subsubsection{Impact of Hyperparameters}
\begin{figure}[h]
\centering
\includegraphics[width=0.7\linewidth]{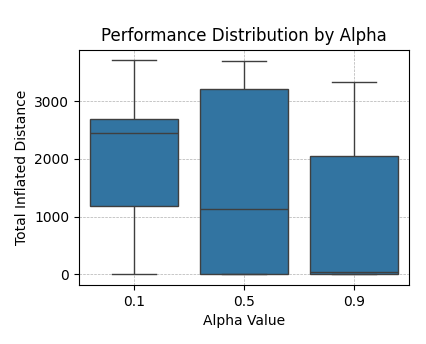}

\caption{Distribution of total inflated distance across all experiments for different values of the spatial weighting parameter $\alpha$ in the spatio--temporal distance metric $D_{ij} = \alpha \tau_{ij} + \beta \Delta T_{ij}$. Lower values of $\alpha$ emphasize temporal proximity, while higher values prioritize spatial proximity. }
\label{fig:alpha}
\end{figure}

\begin{figure}[h]
\centering
\includegraphics[width=0.7\linewidth]{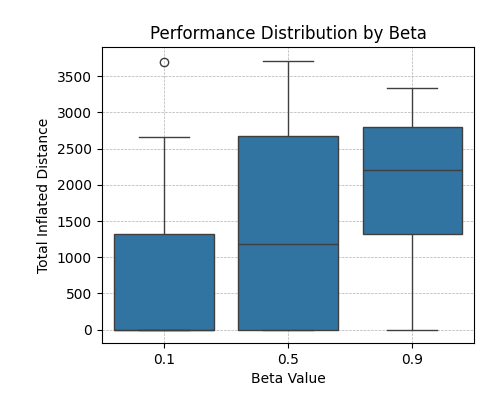}
\caption{Distribution of total inflated distance for different $\beta$ values in the spatio--temporal distance metric. Increasing $\beta$ amplifies the temporal component during node merging as it is the temporal weighting coefficient.}
\label{fig:beta}
\end{figure}

\begin{figure}[h]
\centering
\includegraphics[width=0.7\linewidth]{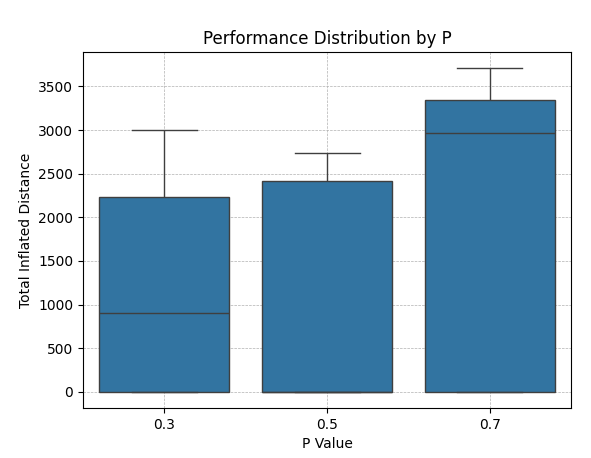}
\caption{Total inflated distance for varying probability thresholds $P$, A higher value of P means that more nodes are considered as potential centers, which could lead to more, but smaller, clusters. A lower value of P will lead to fewer, larger clusters.}
\label{fig:pvalue}
\end{figure}

\begin{figure}[h]
\centering
\includegraphics[width=0.7\linewidth]{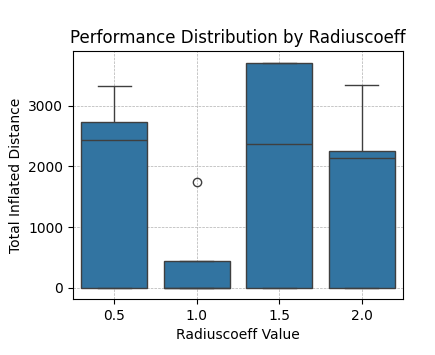}
\caption{Distribution of total inflated distance as a function of the coarsening radius coefficient., which scales the radius that the coarsener uses to find other nodes to group together.}
\label{fig:radius}

\end{figure}

To understand the sensitivity of the coarsening algorithm to its hyperparameters, we performed a random search over 20 trials per instance, guided by the objective score defined in Equation~(1). The results, summarized in Figures~\ref{fig:alpha},~\ref{fig:beta},~\ref{fig:pvalue} and~\ref{fig:radius}, reveal distinct sensitivities for each parameter.

\textbf{Target Node Percentage ($P$):} As shown in Figure~\ref{fig:pvalue}, there is a clear trade-off between coarsening depth and solution quality. Higher values of $P$ (less coarsening, more nodes retained) produce lower and more tightly distributed objective scores, indicating better median quality. Lower values of $P$ increase variance substantially: aggressive coarsening occasionally finds excellent solutions by collapsing clustered customers into well-ordered super-nodes, but more frequently degrades quality by forcing temporally incompatible merges. This suggests that $P = 0.5$ provides the best balance for general use, retaining enough problem structure for the solver to navigate effectively while still delivering meaningful graph reduction.

\textbf{Radius Coefficient ($R_\text{coeff}$):} Figure~\ref{fig:radius} shows a non-monotonic relationship between the radius coefficient and solution quality, consistent with the existence of a sweet spot. Very small values ($R_\text{coeff} = 0.5$) yield negligible coarsening because few candidate pairs fall within the merge radius, providing no benefit. Very large values ($R_\text{coeff} = 2.0$) allow geometrically distant or temporally incompatible nodes to merge, producing poor-quality super-nodes that the inflation step cannot adequately repair. Values near $R_\text{coeff} = 1.5$ produced the most consistent results, and this value is recommended as a default.

\textbf{Spatial and Temporal Weights ($\alpha$, $\beta$):} As shown in Figures~\ref{fig:alpha} and~\ref{fig:beta}, the relative importance of spatial versus temporal proximity depends on the instance family. For clustered (C-type) instances, where customers are geographically concentrated, a higher spatial weight ($\alpha = 0.9$) performs better, since nearby customers are also more likely to be temporally compatible. For random (R-type) instances, where customer locations are dispersed and time windows can span large geographic distances, a balanced weighting ($\alpha = 0.5$, $\beta = 0.5$) yields better outcomes by avoiding merges that are spatially close but temporally incompatible. Practitioners should therefore tune $\alpha$ and $\beta$ based on the known structure of their instance family, rather than using a single fixed setting.

\section{Conclusion}

This paper introduced and experimentally validated a spatio-temporal graph coarsening and refinement strategy for the Capacitated Vehicle Routing Problem with Time Windows (CVRPTW). By combining a composite distance metric $D_{ij} = \alpha \tau_{ij} + \beta \Delta T_{ij}$ with explicit merge-feasibility enforcement and reversible super-node inflation, the proposed pipeline reduces problem scale while preserving the temporal structure of the original instance.

For classical heuristics, the impact of coarsening is strongly solver-dependent. The Greedy heuristic benefits consistently: across 45 Solomon instances, coarsening improved total distance in 91\% of cases, with a median improvement of 36.57\%, a median vehicle count reduction of 33.33\%, and a median computation time reduction of 87.34\%, with no introduction of new constraint violations. The Savings heuristic, by contrast, experiences a systematic deterioration in solution distance after coarsening (median $-33.17\%$ across all instances), because the merge operation distorts the pairwise savings structure on which the algorithm depends. Computation time is nevertheless reduced by a median of 82.44\% for Savings as well, making coarsening a viable strategy for this solver in latency-constrained settings where solution distance is a secondary concern.

For QUBO-based quantum solvers, coarsening provides two complementary benefits whose relative importance depends on instance scale and family. First, it delivers a substantial reduction in computation time: Inflated FQS and APS average $3.3$--$3.4\,\text{s}$ regardless of scale, versus $7.1\,\text{s}$ for Uncoarsened at $N=5$ and $23\,\text{s}$ at $N=10$ — reflecting the quadratic scaling of QUBO variable counts with problem size. Second, on clustered (C-type) instances at $N=10$, coarsening reduces the average vehicle count by $1.00$ and route duration by ${\approx}25\%$ while preserving 100\% feasibility, demonstrating that the reduced search space enables the solver to find structurally more efficient routes. On random (R-type) instances, narrow time windows (median width $= 10$) result in low pairwise temporal overlap, making valid super-node formation difficult regardless of hyperparameter choice; Inflated variants incur infeasibility rates of $35$--$43\%$ at $N=10$. Solution validity is 100\% across all 448 solver-instance pairs, confirming the correctness of the QUBO encoding. Most significantly, the coarsening pipeline enables the quantum solvers to produce feasible solutions on full 100-customer Solomon instances — a scale entirely beyond the reach of the uncoarsened formulation — demonstrating that coarsening is not merely an optimisation accelerator but a scalability-enabling mechanism for quantum approaches to CVRPTW.

In summary, graph coarsening is a practically effective preprocessing strategy for CVRPTW. It reliably accelerates both classical and quantum solvers, meaningfully improves solution quality for greedy-type heuristics and quantum inspired solvers, and enables quantum formulations to operate on problem sizes that would otherwise cause combinatorial explosion.

\section{Future Work}

The consistent degradation observed under the Savings heuristic establishes an important boundary condition: graph coarsening is beneficial for solvers that operate locally on the reduced search space, but counterproductive for algorithms whose optimality depends on global pairwise distance structure. A systematic investigation of which solver architectures are compatible with coarsening as a preprocessor and the structural properties that determine compatibility remains an open and practically relevant direction.

The present evaluation measures improvement relative to (a) the same solver operating without coarsening, and (b) an OR-Tools CP-SAT reference generated for the sub-instance scales $N=5$ and $N=10$. Comparison against best-known Solomon solutions on the full 100-customer scale, or against state-of-the-art exact and meta-heuristic solvers such as LKH-3, is left for future work. The primary contribution of this paper is the coarsening pipeline itself and its ability to extend quantum solvers to otherwise intractable scales, rather than claiming absolute solution optimality.

All experiments use Solomon benchmark instances with Euclidean travel times. The hyperparameter recommendations derived here  particularly the $\alpha$/$\beta$ balance  may require retuning for instances with non-Euclidean travel costs, heterogeneous fleets, or stochastic time windows. Evaluating the coarsening pipeline on real-world logistics datasets with road-network distances is an important direction for future work.

\newpage

\bibliographystyle{IEEEtran}
\bibliography{references}

\section*{Acknowledgment}
This article was prepared as part of the project “Solving routing problems using quantum annealing”, mentored by Paweł Gora and initiated during the QIntern 2025 program organized by the QResearch Department of QWorld. The author expresses sincere gratitude to QWorld and to the project team members for their valuable support and guidance throughout the development of this work.

\appendix

\clearpage

\begin{figure}[htbp]
    \centering
    \subfloat[Baseline (P=1.0, no coarsening)]{%
        \includegraphics[width=0.6\linewidth]{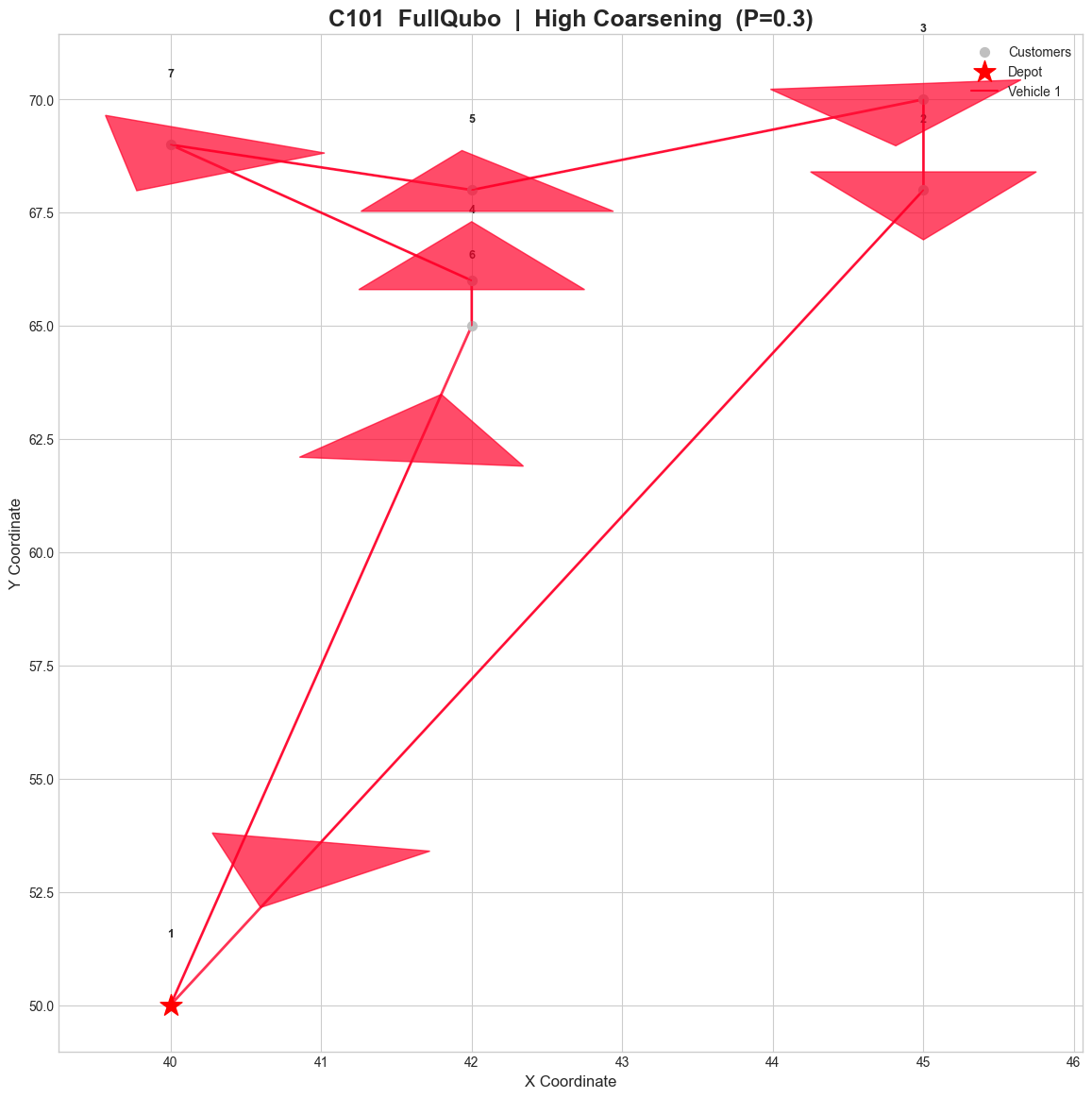}%
        \label{fig:sub1}}
    \\ 
    \subfloat[Low Coarsening (P=0.7)]{%
        \includegraphics[width=0.6\linewidth]{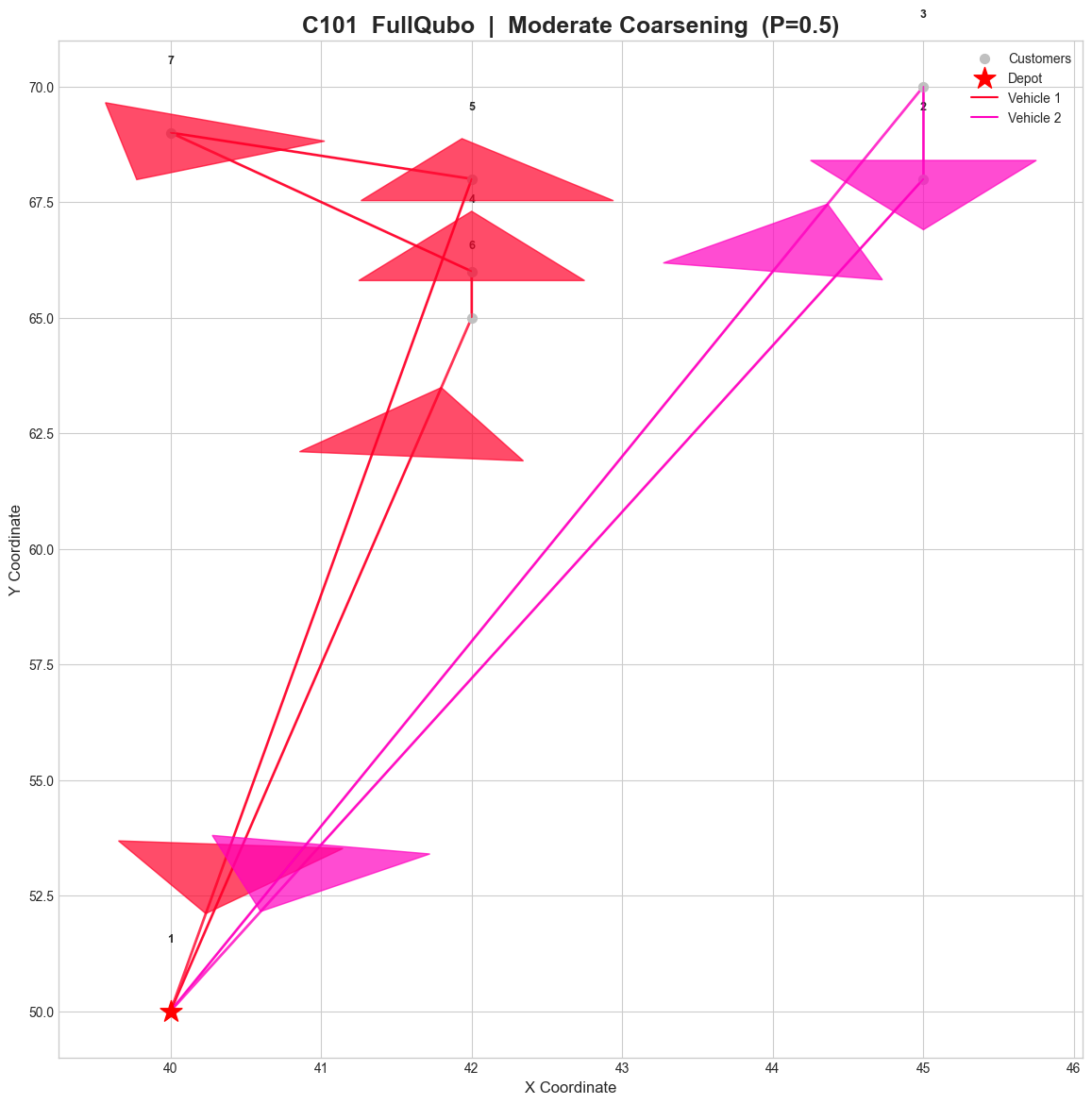}%
        \label{fig:sub2}}
    \\
    \subfloat[Moderate Coarsening (P=0.5)]{%
        \includegraphics[width=0.65\linewidth]{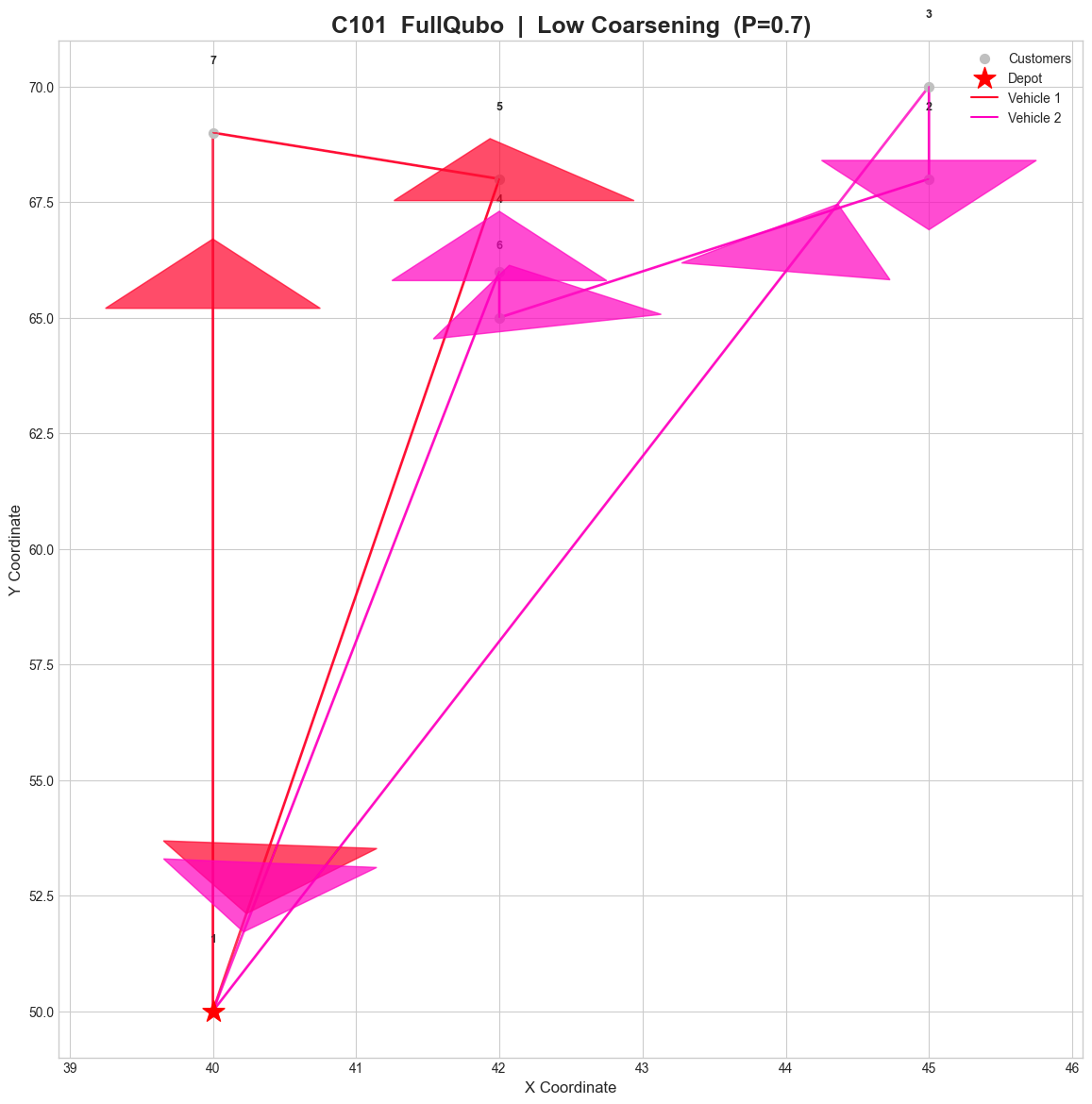}%
        \label{fig:sub3}}
    \\
    \subfloat[High Coarsening (P=0.3)]{%
        \includegraphics[width=0.65\linewidth]{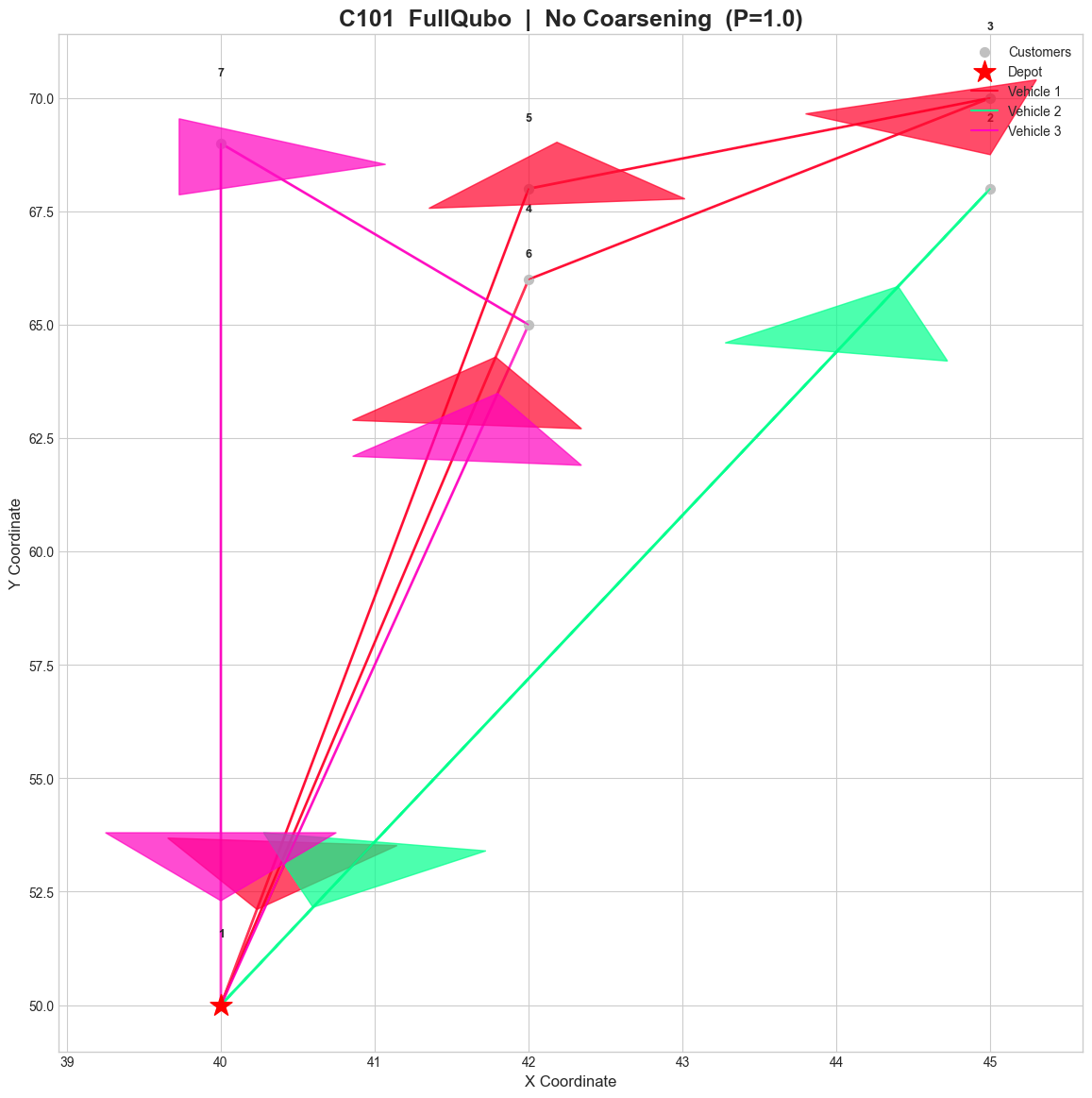}%
        \label{fig:sub4}}
    \caption{Exemplary results of FullQubo solver for C101
instance and different coarsening rate values.}
    \label{fig:vertical_subfig}
\end{figure}

\begin{figure}[htbp]
    \centering
    \subfloat[Baseline (Radius = 1.0)]{%
        \includegraphics[width=0.60\linewidth]{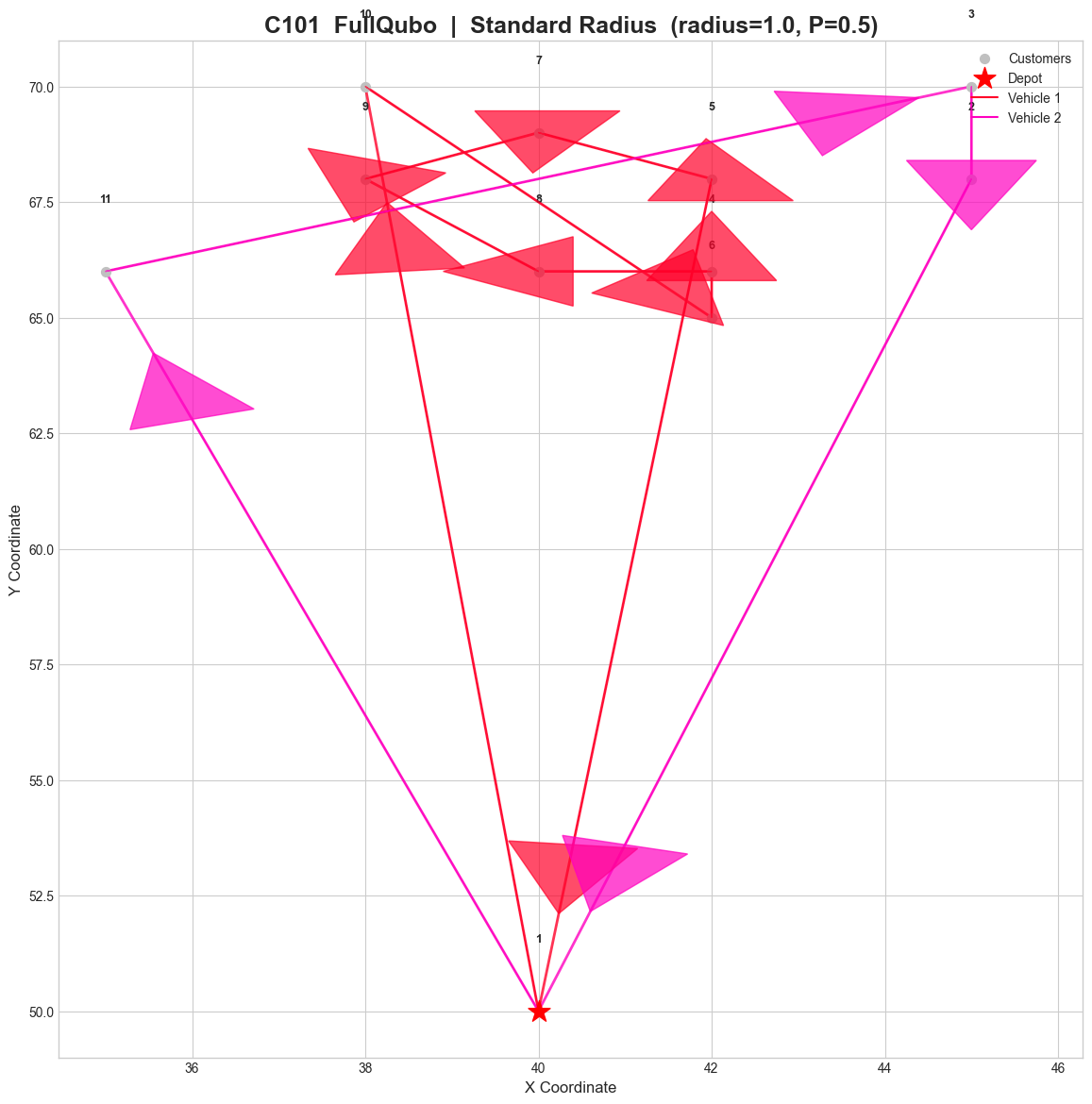}%
        \label{fig:sub11}}
    \\ 
    \subfloat[Low Radius (Radius = 2)]{%
        \includegraphics[width=0.6\linewidth]{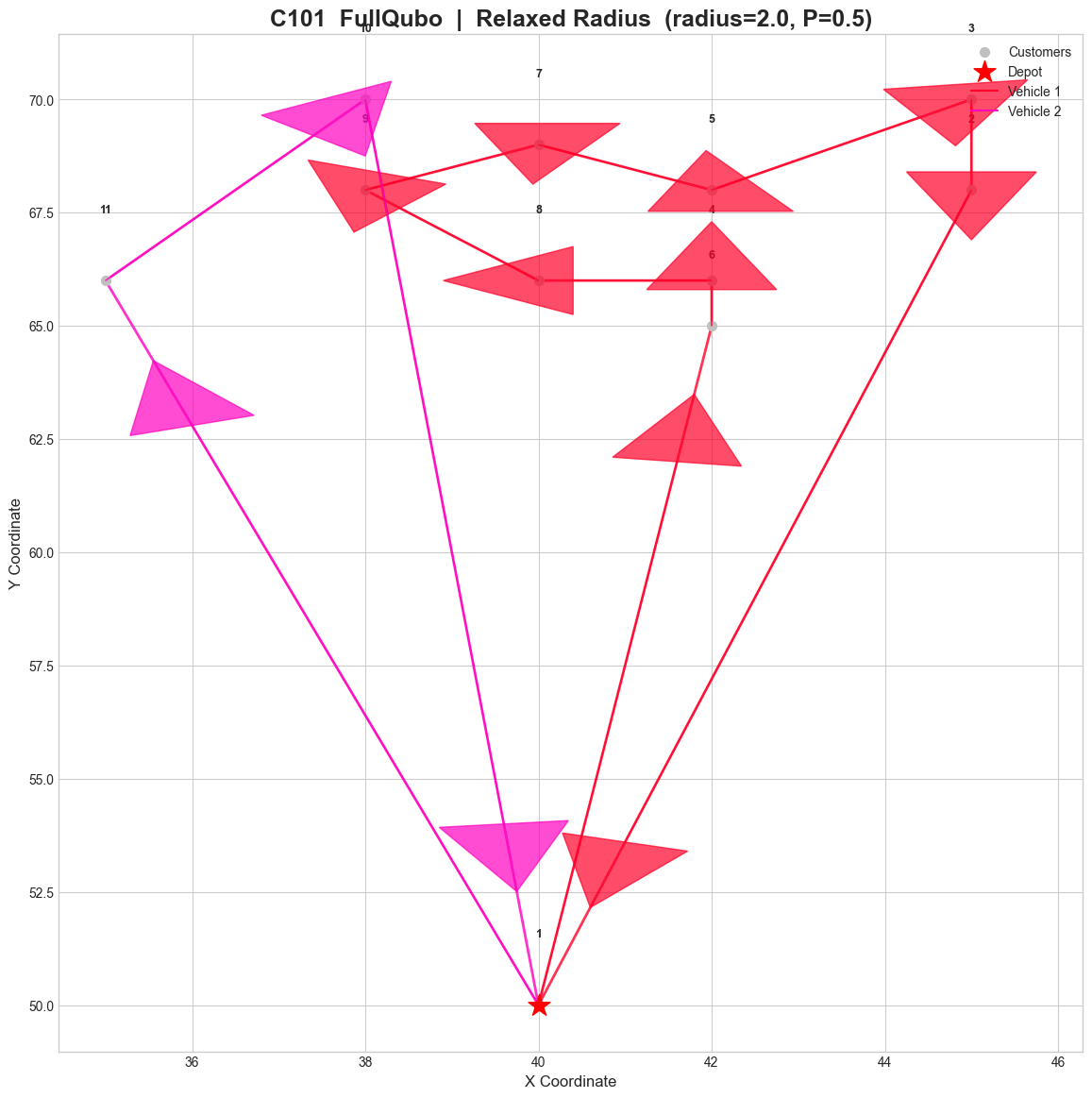}%
        \label{fig:sub22}}
    \\
    \subfloat[Moderate Radius (Radius = 3)]{%
        \includegraphics[width=0.6\linewidth]{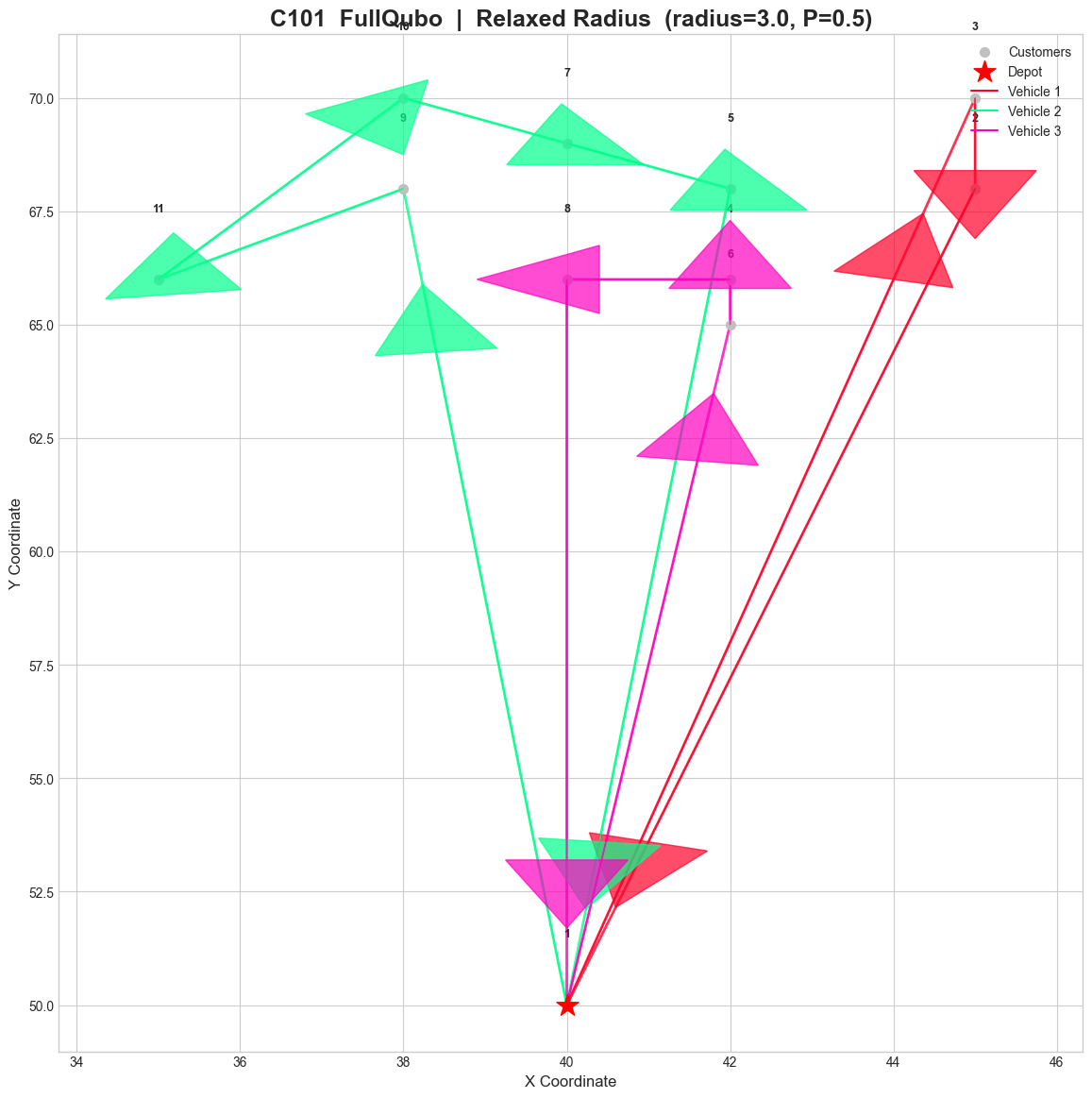}%
        \label{fig:sub33}}
    \\
    \subfloat[High Radius (Radius = 5)]{%
        \includegraphics[width=0.6\linewidth]{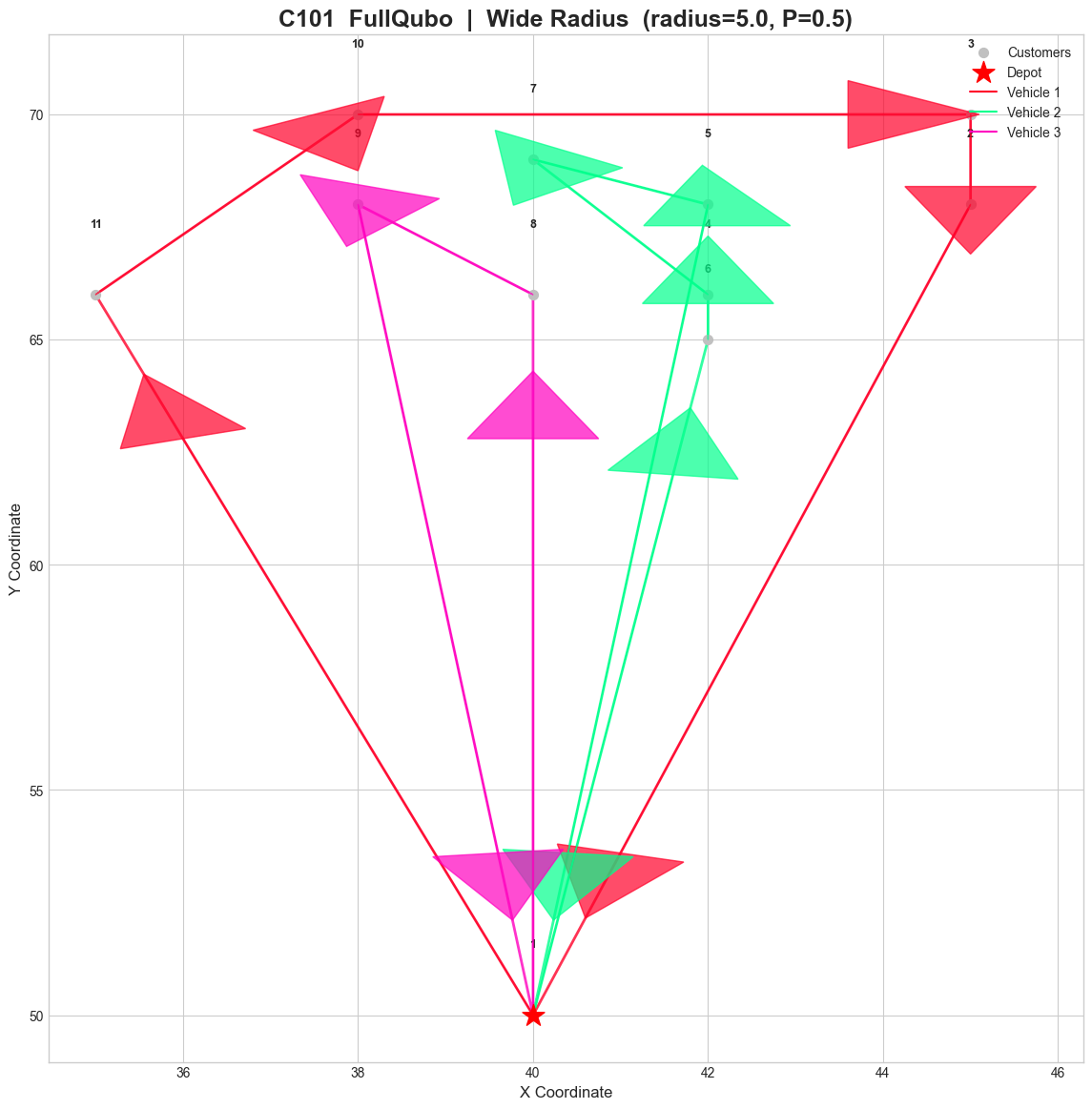}
        \label{fig:sub44}}
    \caption{Exemplary results of Savings solver for C101
instance and different radius values.}
    \label{fig:vertical_subfig2}
\end{figure}

\begin{figure*}[htbp]
    \centering
    \subfloat[Paired Times per Dataset (Savings)\label{fig:paired_savings}]{
        \includegraphics[width=0.35\linewidth]{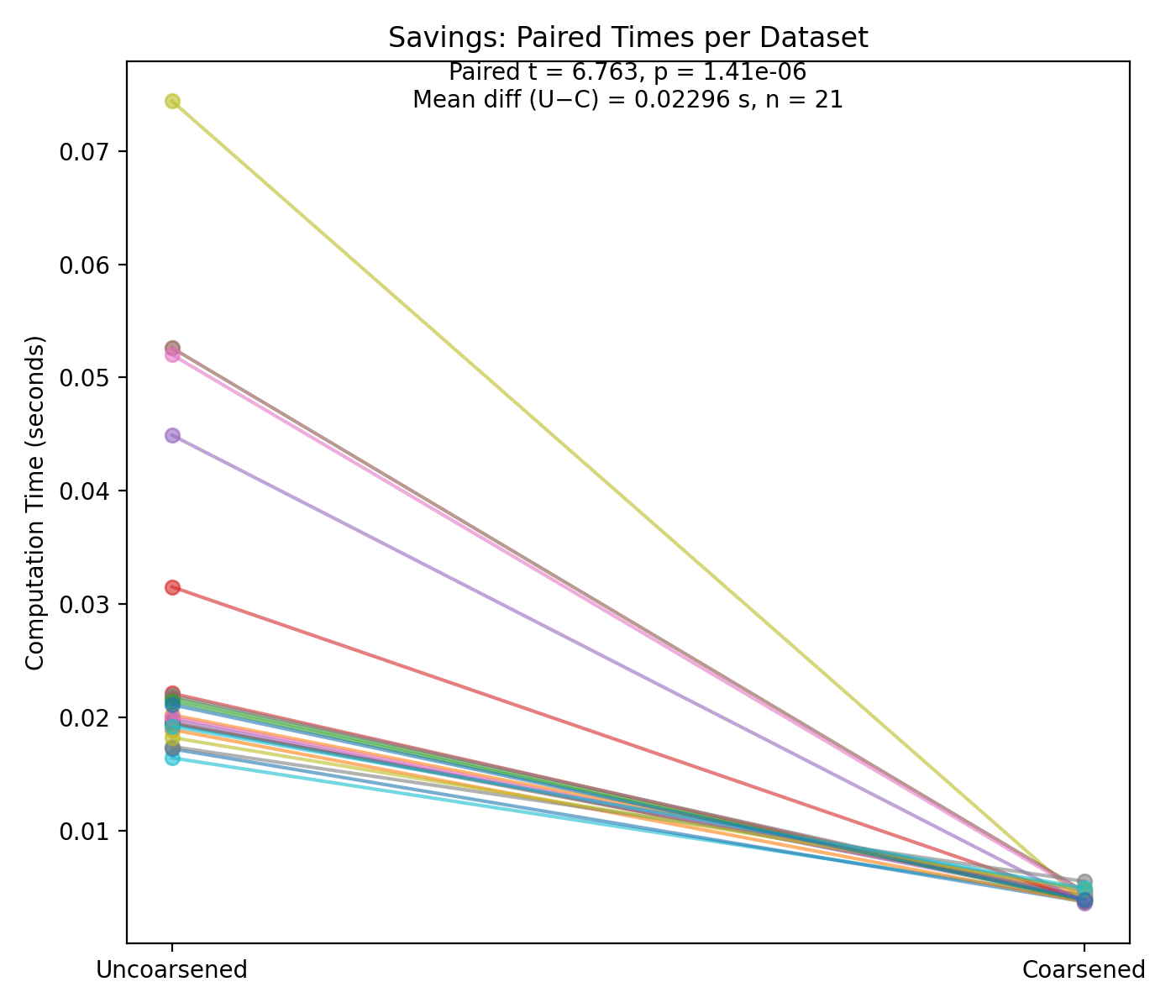}
    }
    \hfill
    \subfloat[Distribution of Paired Differences (Savings)\label{fig:diff_hist_savings}]{
        \includegraphics[width=0.35\linewidth]{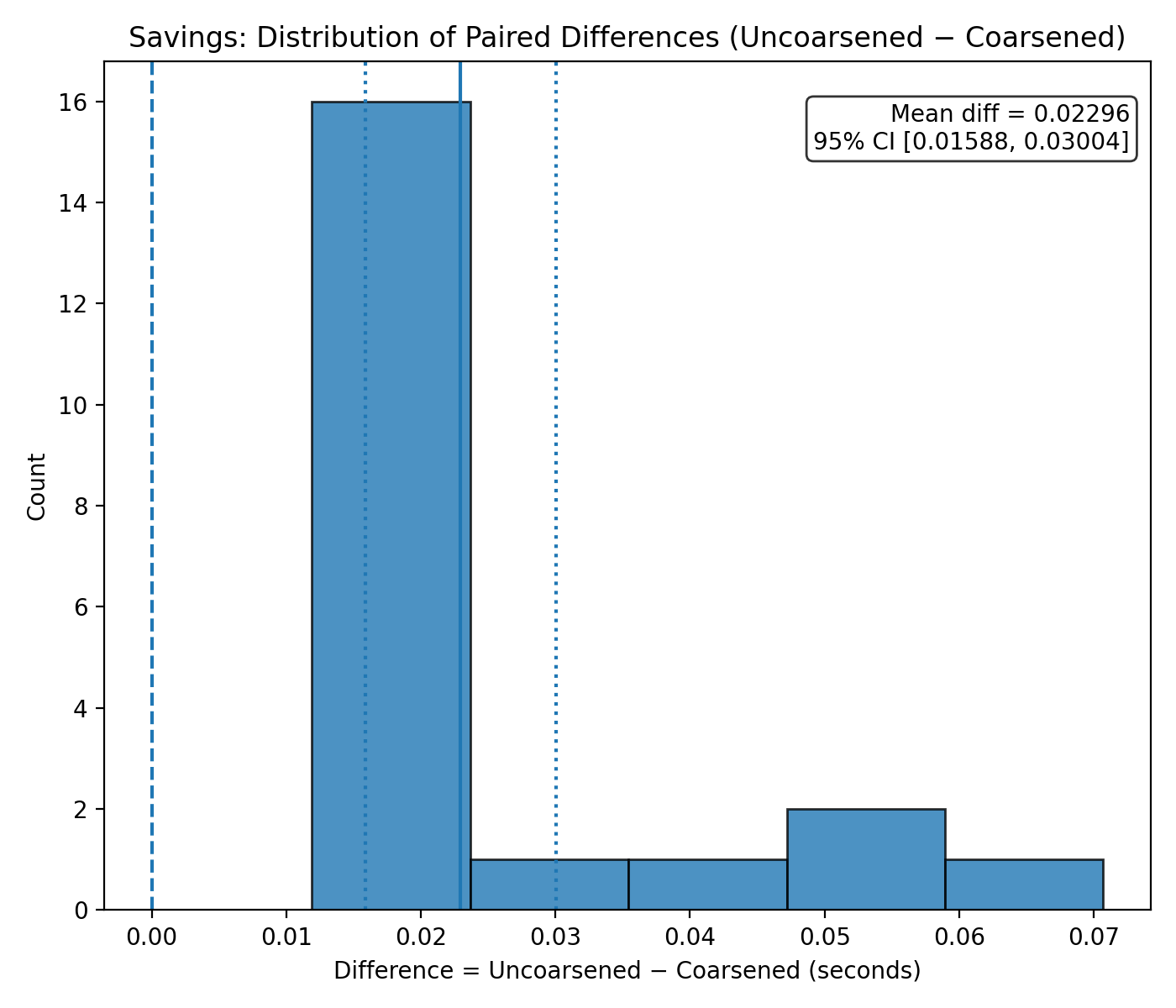}
    }
    \\[-1mm]
    \subfloat[Paired Times per Dataset (Greedy)\label{fig:paired_greedy}]{
        \includegraphics[width=0.35\linewidth]{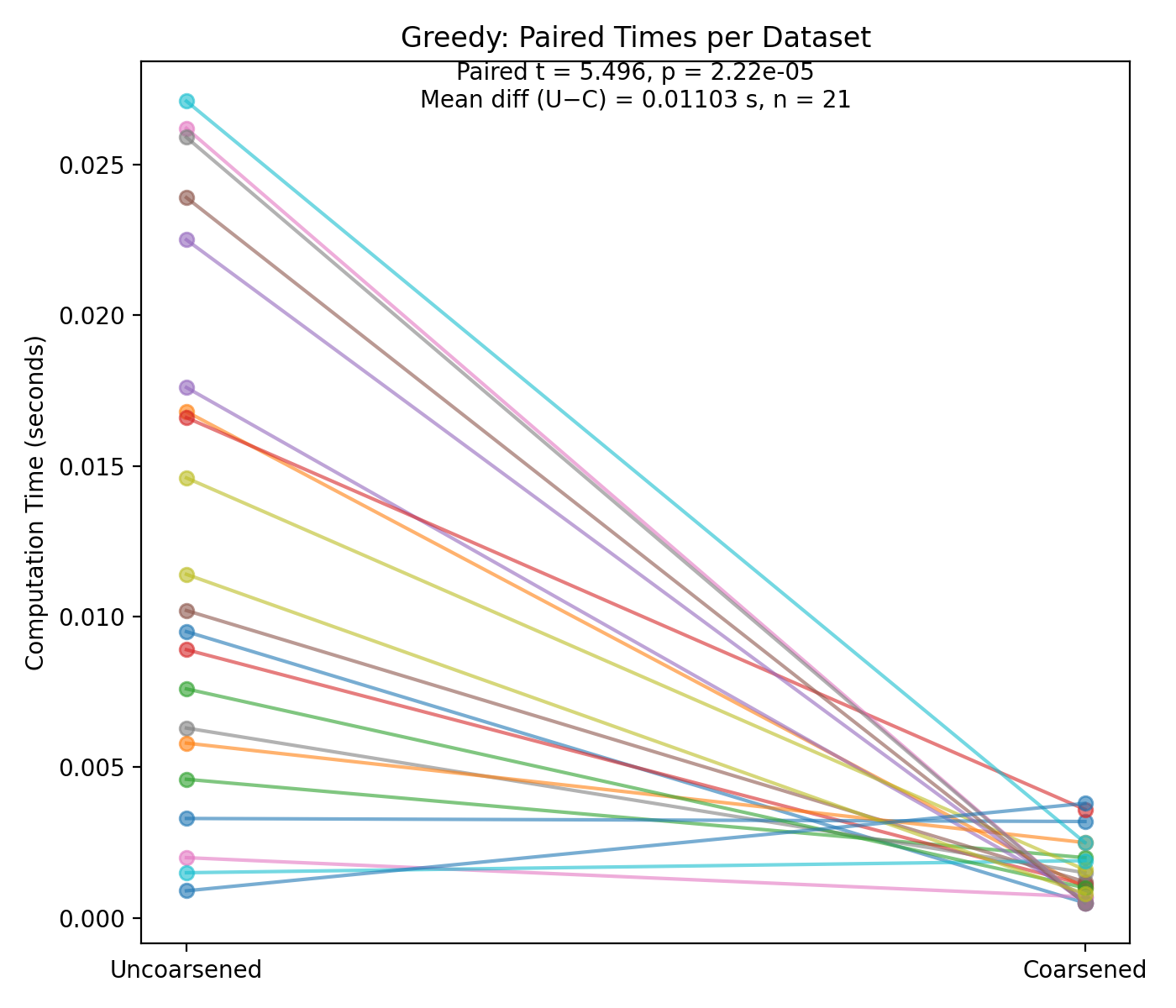}
    }
    \hfill
    \subfloat[Distribution of Paired Differences (Greedy)\label{fig:diff_hist_greedy}]{
        \includegraphics[width=0.35\linewidth]{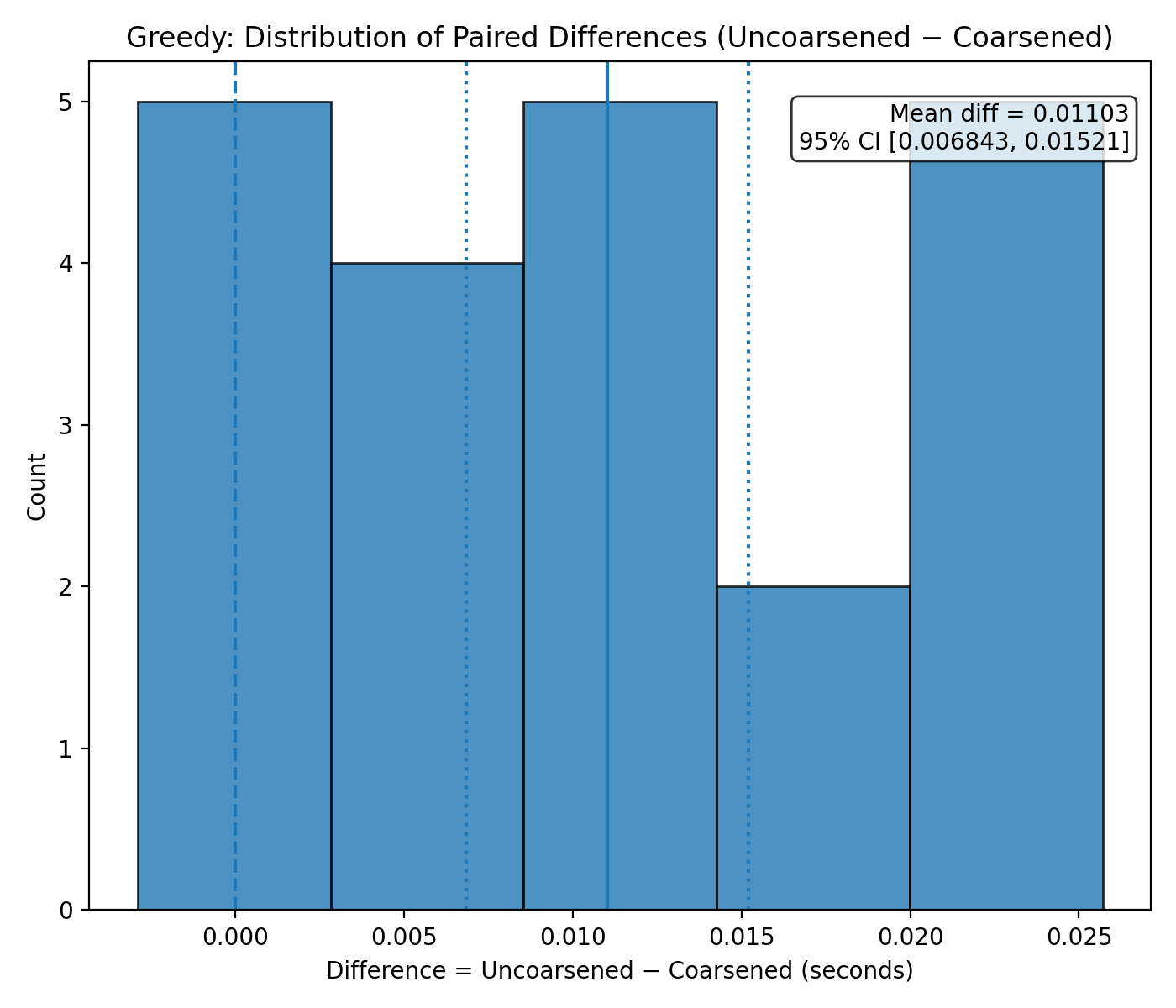}
    }
    \caption{Visualization of paired $t$-test results comparing uncoarsened and coarsened computation times for both Savings and Greedy algorithms.
    (a) and (c) show paired computation times per dataset, where most lines slope downward, indicating faster execution after graph coarsening.
    (b) and (d) depict the distributions of paired differences (Uncoarsened - Coarsened), with positive means and 95\% confidence intervals excluding zero.
    Together, these results show that graph coarsening significantly reduced computation time for both algorithms, with a stronger effect for the Savings method.}
    \label{fig:ttest_results}
\end{figure*}

\begin{figure}[htbp]
    \centering
    \includegraphics[width=0.60\textwidth]{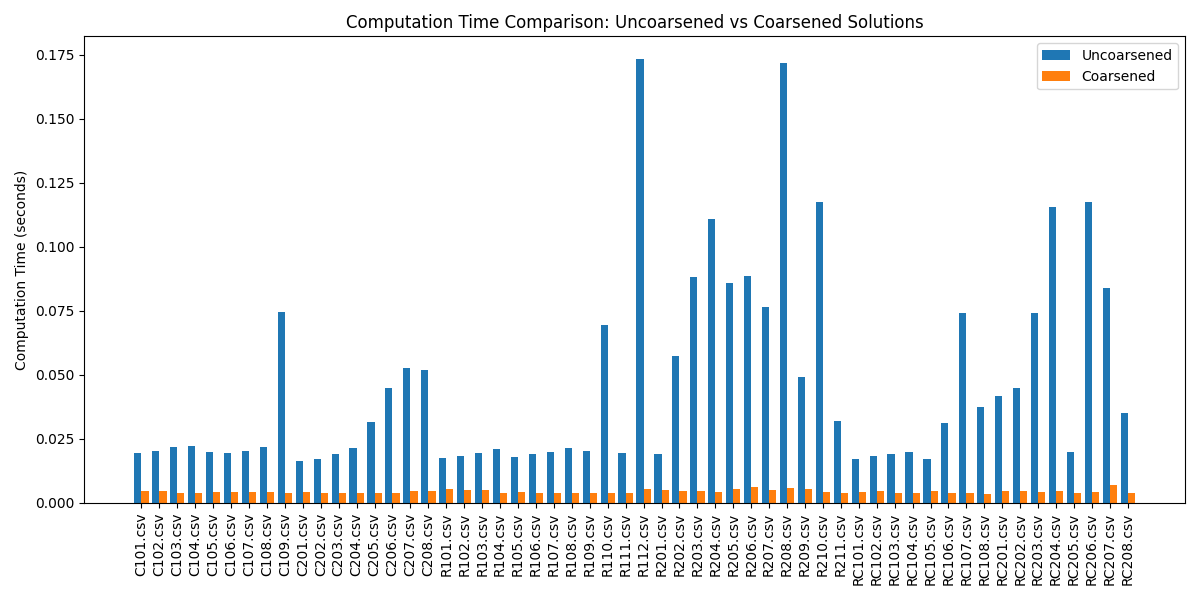}
    \caption{Average computation time comparison between Uncoarsened and Coarsened methods 
    for both Savings and Greedy solutions.}
    \label{fig:avg_computation_time_overall}
\end{figure}

\end{document}